\documentclass{article}
\usepackage{arxiv}

\usepackage[utf8]{inputenc} 
\usepackage[T1]{fontenc}    
\usepackage{hyperref}       
\usepackage{url}            
\usepackage{booktabs}       
\usepackage{amsfonts}       
\usepackage{nicefrac}       
\usepackage{microtype}      

\usepackage{graphicx}

\usepackage{esvect}
\usepackage{amsmath}
\usepackage[tight,footnotesize]{subfigure}
\usepackage{appendix}
\usepackage{lscape}
\usepackage{textcomp}
\usepackage{amssymb}
\usepackage{lscape}
\usepackage[noend]{algpseudocode}
\usepackage[boxruled]{algorithm2e}
\usepackage{hhline}
\usepackage{slashbox}
\usepackage{multirow}
\usepackage{lettrine}
\usepackage{tabularx}
\usepackage{makecell}

\title{Parkinson's Disease Diagnosis based on Gait Cycle Analysis Through an Interpretable Interval Type-2 Neuro-Fuzzy System}

\author{
  Armin Salimi-Badr \\
  Faculty of Computer Science and Engineering\\
  Shahid Beheshti University\\
  Tehran, Iran \\
  \texttt{a\_salimibadr@sbu.ac.ir} \\
  \And
   Mohammad Hashemi \\
  Faculty of Computer Science and Engineering\\
  Shahid Beheshti University\\
  Tehran, Iran \\
  \texttt{mohamm.hashemi@mail.sbu.ac.ir} \\
  \And
    Hamidreza Saffari \\
  Faculty of Computer Science and Engineering\\
  Shahid Beheshti University\\
  Tehran, Iran \\
  \texttt{h.saffari@mail.sbu.ac.ir} \\
}

\begin{document}
\maketitle

\begin{abstract}
In this paper, an interpretable classifier using an interval type-2 fuzzy neural network for detecting patients suffering from Parkinson's Disease (PD) based on analyzing the gait cycle is presented. The proposed method utilizes clinical features extracted from the vertical Ground Reaction Force (vGRF), measured by 16 wearable sensors placed in the soles of subjects' shoes and learns interpretable fuzzy rules. Therefore, experts can verify the decision made by the proposed method based on investigating the firing strength of interpretable fuzzy rules. Moreover, experts can utilize the extracted fuzzy rules for patient diagnosing or adjust them based on their knowledge. To improve the robustness of the proposed method against uncertainty and noisy sensor measurements, Interval Type-2 Fuzzy Logic is applied. To learn fuzzy rules, two paradigms are proposed: 1- A batch learning approach based on clustering available samples is applied to extract initial fuzzy rules, 2- A complementary online learning is proposed to improve the rule base encountering new labeled samples. The performance of the method is evaluated for classifying patients and healthy subjects in different conditions including the presence of noise or observing new instances. Moreover, the performance of the model is compared to some previous supervised and unsupervised machine learning approaches. The final Accuracy, Precision, Recall, and F1 Score of the proposed method are 88.74\%, 89.41\%, 95.10\%, and 92.16\%. Finally, the extracted fuzzy sets for each feature are reported.\\

\end{abstract}

\section{Introduction}

Emerging symptoms of some chronic diseases like Parkinson's Disease are diagnosed many years after infection. This delay reduces the chance of encountering the disease with some more simple treatment methods. Parkinson’s disease (PD) is the second most common neurological disease that puts the lives of a large portion of the elderly population in danger \cite{khoury2019data,hariharan2014new,pan2012parkinson}. This slow and progressive neurodegenerative disorder is the consequence of destroying dopaminergic neurons in the Substantia Nigra pars compacta, one nucleus of a complex subcortical brain structure named Basal Ganglia \cite{guyton2010,salimi2017possible,salimi2018system,khoury2019data}.   According to the Parkinson's disease foundation, Most PD patients show movement disorders including movement slowness (Bradykinesia), impairment of the power of voluntary movements (Akinesia), rigidity, and resting tremor \cite{guyton2010}. Near 5 million people are affected by PD in the world \cite{khoury2019data}. The age-standardized mortality rate in Iran is reported about $3.44\%$ \cite{fereshtehnejad2015mortality}.

It would be possible to control and decrease the syndromes by medical treatments or changing lifestyles in the early stages of the disease. However, by progressing this illness, more complex and aggressive approaches including surgical therapy like using Deep Brain Stimulus (DBP) are required. Unfortunately, the PD begins 5 to 10 years before appearing any clinical symptoms \cite{khoury2019data}. Moreover, some non-typical signs such as depression, pain, fatigue, etc. could make the diagnosis more difficult. Generally, the diagnosis task is difficult based on the reported misdiagnosis rate of $25\%$ of cases \cite{das2010comparison}. Machine learning, ambient intelligence, and context-aware agents can help us to improve the quality of healthcare services and disease diagnosis.

Two important symptoms, studied in the previous methods to classify the patients and healthy persons are \textit{speech disorders (vocal impairment)} \cite{das2010comparison,aastrom2011parallel,hariharan2014new}and \textit{gait cycle changing} \cite{khoury2019data,khoury2018cdtw,salimipd}. Since PD affects mainly the patient's movements by disorders like rigidity, Bradykinesia, Akinesia, tremor, and difficulty with balance and coordination its effects on the gait cycle are expected.

Most previous approaches that applied machine learning paradigms for classifying patients suffering from PD have not used either an interpretable artificial intelligence method \cite{khoury2019data,khoury2018cdtw} or they not used expert understandable clinical features \cite{salimipd,36lee2012parkinson,qbp}. However, in a sensitive clinical application, using an interpretable method with clinical features is more acceptable for two reasons: 1- An expert can verify the decision of the model, and 2- an expert can modify and fine-tune the parameters of the model.

In this paper, an interpretable artificial intelligence-based approach using the clinical features extracted from the gait cycle is proposed. The paradigm is based on training an interpretable interval type-2 fuzzy neural network and applying a hybrid batch-online learning method. First, high-level features explaining the behavior of the gait cycle obtained from wearable sensors are extracted. Next, using available training instances, an interval type-2 fuzzy neural network with interpretable structure is trained. Based on training this interpretable structure, some fuzzy rules are extracted which can be used as a guide for experts to diagnose patients. On the other hand, experts can modify and fine-tune the parameters of extracted fuzzy rules based on their knowledge.

Moreover, sensors measure noise along with the target signal. Therefore, using the interval type-2 fuzzy logic increases the robustness of the extracted fuzzy rules against the noisy data. Considering the progress of studies and collecting more labeled instances in the future, a complementary online learning method is proposed to add new fuzzy rules encountering new training samples. The main contributions of our proposed method are summarized as follows:

\begin{enumerate}
  \item Presenting an interpretable model based on clinical features. Therefore, the decisions made by the model can be verified by the experts. Moreover, extracted rules can be used by experts or be adjusted based on the experts' knowledge;
  \item Increasing the robustness of the method against uncertainty and sensor noisy measurements using Interval Type-2 Fuzzy logic;
  \item Proposing an initial batch learning to extract fuzzy rules based on available training samples and proposing complementary online learning for improving the model's rule base using new labeled instances;
  \item Avoiding local search like Gradient Descent for tuning network's parameters. Using local search optimizations could lead overfitting.
\end{enumerate}

The rest of this paper is organized as follows: First, in section \ref{sec3} the related studies are reviewed. Afterward, in section \ref{sec4} the proposed paradigm including the utilized features, preprocessing, the proposed Interval Type-2 Fuzzy Neural Network architecture, and learning algorithms are explained. Next, the results on some labeled data are reported and compared with some other methods that used similar clinical features in section \ref{sec5}. At the end of section \ref{sec5}, the extracted fuzzy sets are reported. Finally, conclusions are presented in section \ref{sec6}.

\section{Related Work}
\label{sec3}

The deterioration of executive functions and movement disorders in patients with PD have been shown extensively \cite{litvan2012diagnostic,marras2013measuring,baiano2020prevalence}. Yogev et al. \cite{yogev2005dual} studied the impacts of different types of dual-tasking and cognitive function on the gait of patients with PD and control subjects. They also showed contrasting measures of gait rhythmicity for patients with PD in comparison to other features. Additionally, in \cite{yogev2007gait} it is investigated that Parkinson's disease has a great impact on the left-right symmetry of gait. Yogev et al. \cite{yogev2007gait} conducted a similar walking condition for both patients with PD and healthy control subjects and they demonstrated that asymmetry of gait increased mainly during the dual-task condition in patients with PD but not in the healthy control subjects. Considering the Hausdorff et al. \cite{hausdorff1998gait} studies on gait variability and Basal Ganglia disorders, it can be concluded that the ability to maintain a steady gait with low stride-to-stride variability of gait cycle timing, would be decreased in patients with PD. Parkinson's disease symptoms also include speech disorders as well as cognitive impairments \cite{pahwa2013handbook,harel2004variability}. In addition, $90\%$ of the patients with PD themselves report speech impairments as one of the most significant symptoms \cite{hartelius1994speech}.

To classify the patients with PD and healthy control subjects based on their gait cycles, both wearable \cite{10jeon2008classification,11ashhar2017wearable,12nieuwboer2004electromyographic,14hong2009kinematic,15saito2004lifecorder,17salarian2004gait,18mariani2012shoe} and non-wearable \cite{19latash1995anticipatory,20cho2009vision,21pachoulakis2014building,22galna2014accuracy,23dror2014automatic,24dyshel2015quantifying,25antonio2015abnormal,26song2012altered,27foreman2012improved,28muniz2010comparison,29vaugoyeau2003coordination} sensors have been used in various experiments. For instance, Jean et al. \cite{10jeon2008classification}, conducted the classification using Spatial-Temporal Image of Plantar pressure (STIP) among normal step and patient steps with PD. In \cite{18mariani2012shoe}  Mariani et al. presented a technology based on wearable sensors on-shoe and utilizing algorithms, which had an outcome of characterizing Parkinson's disease motor symptoms during ‘Timed Up and Go’ (TUG) and gait tests.

Accordingly, the classification of healthy control subjects and patients with PD using ground reaction force sensors placed in shoes has been extensively studied \cite{30su2015characterizing,31zeng2016parkinson,32daliri2012automatic,36lee2012parkinson,salimipd,khoury2018cdtw,khoury2019data}. In \cite{32daliri2012automatic}, the vGRFs measurements of both left and right foot were used to extract statistical features including minimum, maximum, average, and the standard deviation of each time series. Their extracted features then were fed to machine learning binary classifier including Support Vector Machine (SVM). In \cite{30su2015characterizing} Su et al., the gait asymmetry (GA) was calculated by comparing the ground reaction force (GRF) features of the left and right limbs. This was done by decomposition of the GRF into components of different frequency sub-bands via the wavelet transform and Multi-Layer Perceptron (MLP) models.  In \cite{36lee2012parkinson}, Lee et al. utilized the gait characteristics of idiopathic PD patients who shuffle their feet while they are walking to classify patients with PD and healthy control subjects. They trained a neural network with weighted fuzzy membership functions (NEWFM) using extracted 40 statistical and wavelet-based features.

Different supervised and unsupervised methods such as Decision Tree (DT), Support Vector Machine (SVM), K-Nearest Neighbors (KNN), Gaussian-Mixture Model (GMM), and K-Means are extensively utilized to classify patients with PD and healthy control subjects. In \cite{37joshi2017automatic}, automatic noninvasive identification of PD is used with the combination of wavelet analysis and SVM which led to an accuracy of $90.32\%$. Although most of the previous studies are mainly based on the use of time-domain and frequency-domain features, in \cite{khoury2019data}, only the clinical-based features extracted from vertical Ground Reaction Forces (vGRFs) were considered. Accordingly, nineteen statistical features measuring different aspects of the gait cycle are extracted and used as the input of machine learning-based classifiers. However, most of the previous works tried to extract feature vectors based on some human knowledge, in \cite{salimipd}, demonstrated that the spatial correlation among different sensors data during time placed in each left and right foot are useful for diagnosing the patients. To consider the temporal dependencies, they proposed a structure that has Long-Short Term Memory (LSTM) cell layers to build a Recurrent Neural Network (RNN).

A  time-delay neural network classifier learned by a Q-back propagation learning approach was proposed in \cite{qbp} using temporal information of vGRF time-series to predict the severity of gait disturbances in subjects with PD. The dataset used includes data from three PD research studies \cite{yogev2005dual,frenkel2005treadmill,hausdorff2007rhythmic}. The results show that the classification accuracy on the three sub-datasets reached $91.49\%$, $92.19\%$, and $90.91\%$, respectively.

Most previous studies have not used clinical features or have not applied an interpretable method for classifying PD patients. Using an interpretable machine learning approach that the base of its decision making is transparent and expert understandable is more agreeable in a clinical application. Fuzzy Neural Networks are interpretable structures with neural representation \cite{ANFIS,SOFMLS,Ebadzadeh15,ebadzadeh2017, Salimi2020novel,Salimi-Badr2017,salimi2020backpropagation,salimi2021it2cfnn}. In this study, the clinical features which are understandable for experts are extracted from the recorded vGRF signal and an interpretable structure, based on Interval Type-2 Fuzzy Neural Network is utilized to classify subjects using these clinical features. Consequently, contrary to the previous studies that used clinical features with non-interpretable machine learning methods like \cite{khoury2019data}, or used sequence of vGRF like \cite{36lee2012parkinson}, the proposed method is able to extract expert understandable rules. These rules can be verified and modified based on the knowledge of experts. On the other hand, experts can utilize the extracted rules to detect patients suffering from PD.

\section{Materials and Methods}
\label{sec4}
\begin{figure}[!t]
\centering
\includegraphics[width = 4.5in]{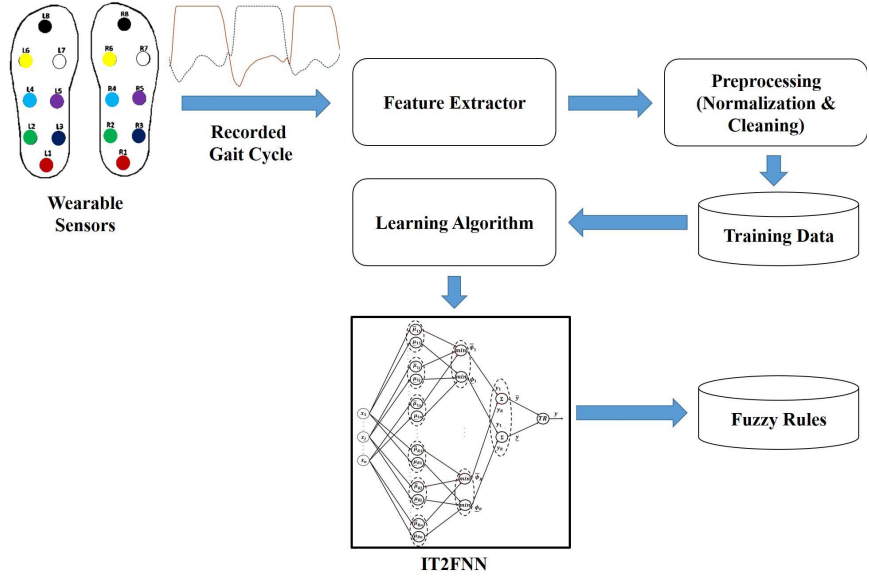}
\caption{The proposed paradigm in a single view. First, the gait cycle in the form of vertical Ground Reaction Force (vGRF) is extracted from some wearable sensors during subject's walking. Afterward, the recorded vGRF signals are preprocessed. Next, clinical features are extracted from the recorded vGRF signals and the training samples are constructed. Using the training samples, an Interval Type-2 Fuzzy Neural Network is trained based on a batch learning algorithm and some fuzzy rules are extracted.}
\label{fig_all}
\end{figure}

In this section, we explain the proposed method including the preprocessing, the architecture of the Interval type-2 Fuzzy Neural Network, the algorithm to learn its parameters, and finally a complementary online learning for improving the method encountering new labeled examples. Figure \ref{fig_all} shows the whole proposed paradigm of this article. In the following subsections, different parts of this paradigm will be explained.

\subsection{Features and Preprocessing}
\label{sec41}
\begin{figure}[t]
\centering
\begin{tabular}{c}
    \subfigure[][]{\includegraphics[width = 4in]{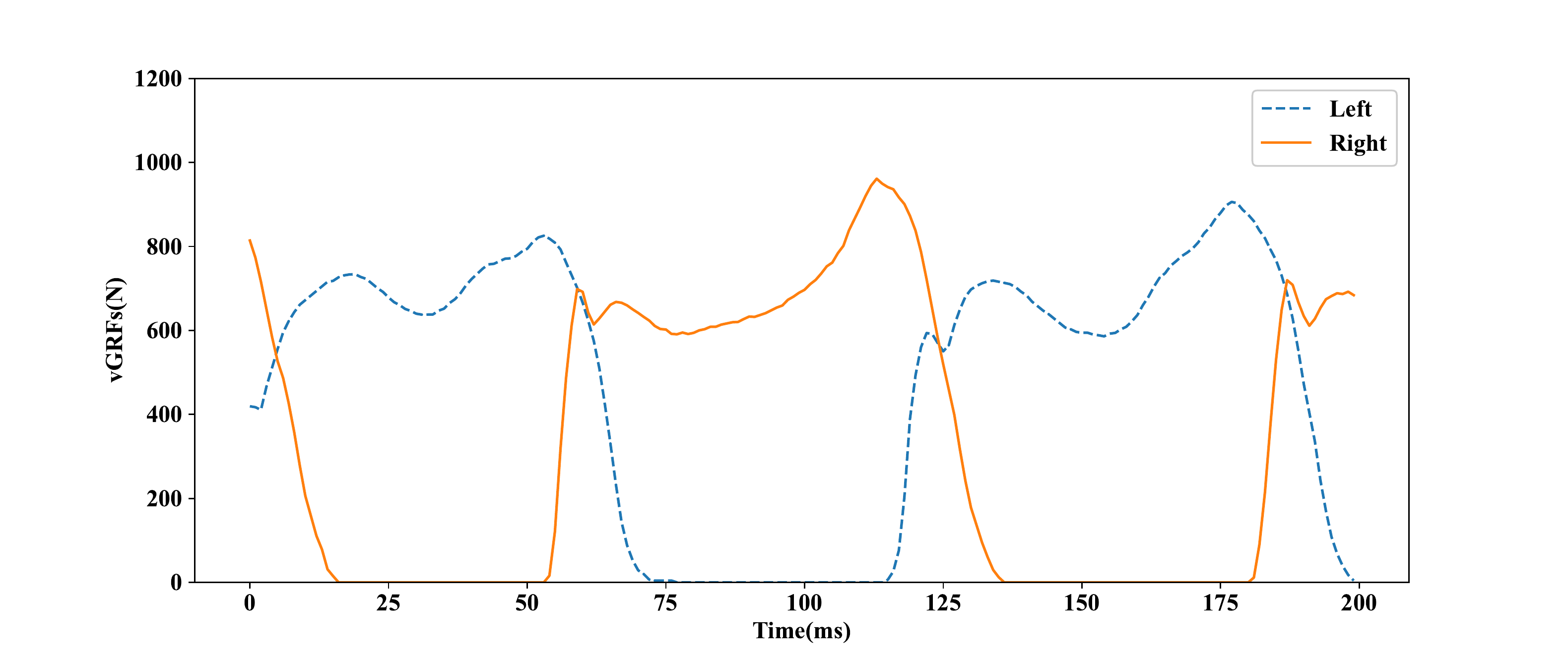}} \\
    \subfigure[][]{\includegraphics[width = 4in]{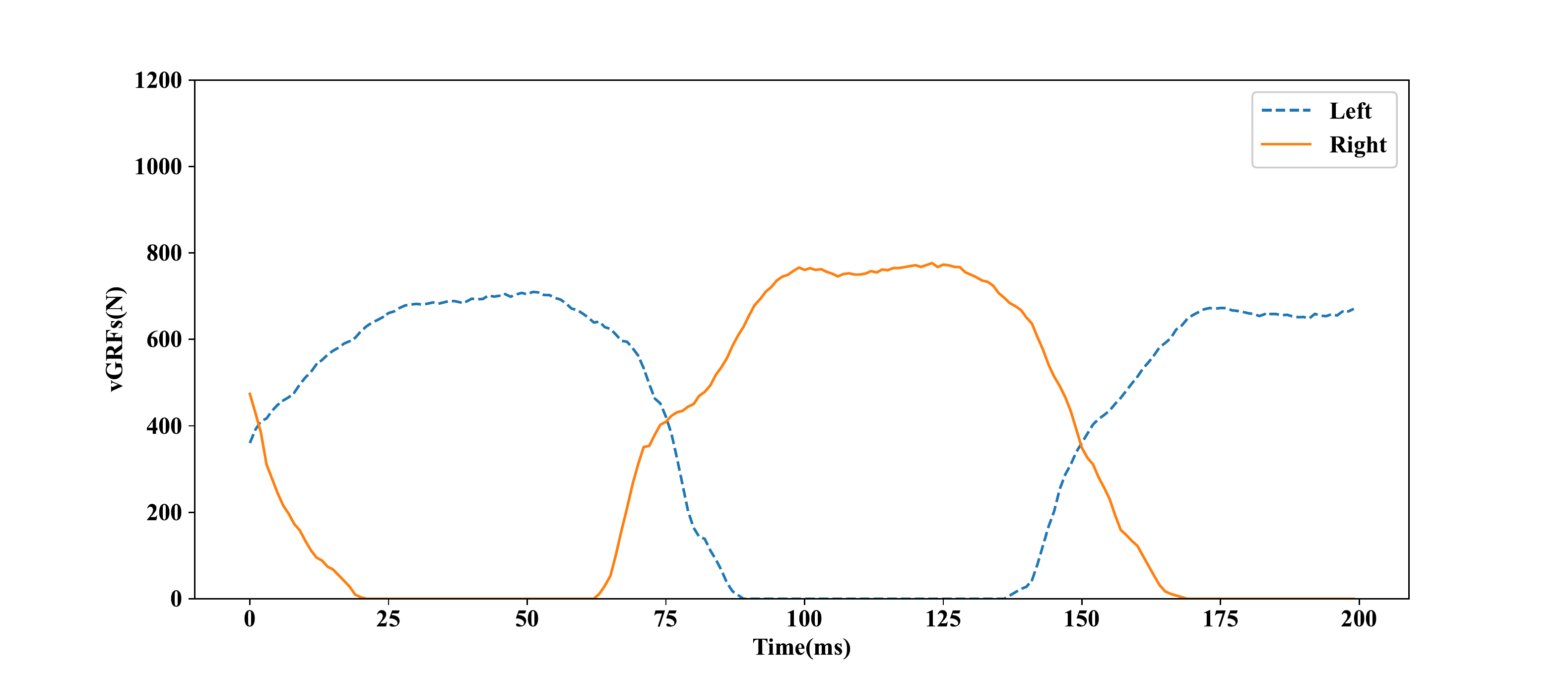}}
\end{tabular}
\caption{Comparing the gait cycle obtained from vGRF of a healthy subject (a) and a patient suffering from Parkinson's Disease (b).}
\label{fig_vgrf}
\end{figure}

\begin{figure}[!t]
\centering
\includegraphics[width = 4in]{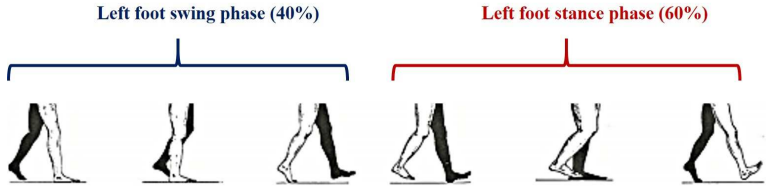}
\caption{Gait cycle (inspired by \cite{khoury2019data,bazner2000assessment}).}
\label{fig0}
\end{figure}

\begin{figure}[!t]
\centering
\includegraphics[width = 1.25in]{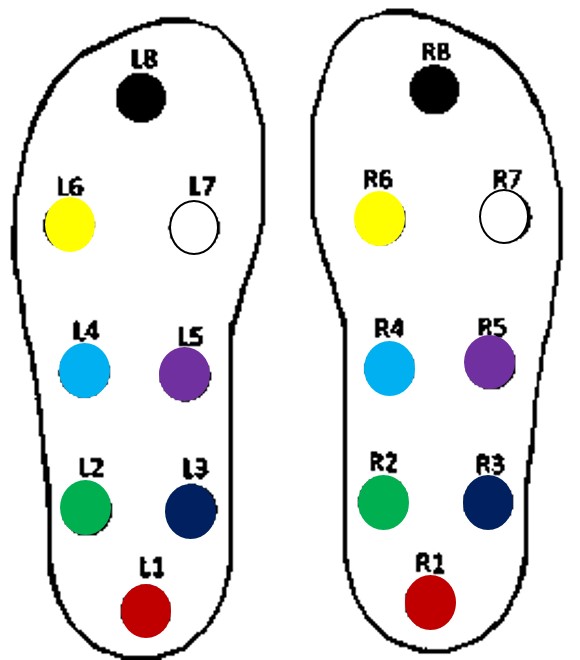}
\caption{Placement of different sensors under feet (inspired by \cite{khoury2019data}).}
\label{fig1}
\end{figure}

The gait cycle is composed of two phases (\ref{fig0}) \cite{khoury2019data,khoury2018cdtw}: stance phase ($60\%$ of the gait cycle) and swing phase ($40\%$ of the gait cycle). Although, healthy persons walking is characterized by a repetition of the gait cycle pattern, significant variations could be seen between different gait cycles during patients' walking (see Figure \ref{fig_vgrf})\cite{khoury2018cdtw}.  Therefore, in this study, the gait cycle is analyzed to classify patients and healthy persons. To extract the gait cycle pattern, 8 sensors are placed in each shoe sole to record vertical Ground Reaction Force (vGRF) during walking (see Figure \ref{fig1}). Afterward, to verify and fine-tune the function of the proposed model, some clinical features are extracted from the recorded vGRF. These high-level clinical features are introduced in Table \ref{table_features}. To reduce the dimension of input variables and considering the correlations among features obtained form either left or right foot, here we average values related to both feet and construct one variable for each foot. The final clinical features are introduced in Table \ref{table_features2}

\begin{table}[!b]\label{table_features}
\tiny
    \caption[c]{List of the fourteen extracted features.}
    \centering
    \begin{tabular}{ccccccccc}
        \hline
        \\
        \textbf{Feature References}  & \textbf{Extracted Features} & \textbf{Explanation} \\
        \\
        \hline
        \\
        $z_1$ & \makecell{Mean of the \\ Short Swing Time}  & \emph{\makecell{compare Swing Time between the Right and the Left foot, then distribute it according \\ to the Short Swing Time Criteria (Mean of the Short Swing Time) \cite{yogev2007gait}}}\\
        \\
        \hline
        \\
		$z_2$ & \makecell{Mean of the \\ Long Swing Time}  & \emph{\makecell{compare Swing Time between the Right and the Left foot, then distribute it according \\ to the Long Swing Time Criteria (Mean of the Long Swing Time) \cite{yogev2007gait}}}\\
		\\
		\hline
		\\
		$z_3$ & \makecell{Mean of the \\ Gait Asymmetry}  & $ 100 \times |ln (\frac{\emph{Short Swing Time}}{\emph{Long Swing Time}})| $ \emph{(averaged of the Gait Asymmetry) \cite{yogev2007gait}}\\
		\\
		\hline
		\\
		$z_4$ & \makecell{Mean in Percentage (\%) \\ of the Swing Time \\ of the Left Foot}  & \makecell{$ 100 \times \frac{\emph{Duration (s) of Swing Time of the left foot}}{\emph{Duration (s) of Stride Time of the left foot} } $ \\ \emph{(Mean of the Percentage (\%) of the Swing Time of the left foot) \cite{yogev2005dual,hausdorff1998gait}}}\\
		\\
		\hline
		\\
		$z_5$ & \makecell{Mean in Percentage (\%) \\ of the Swing Time \\ of the Right Foot}  & \makecell{$ 100 \times \frac{\emph{Duration (s) of Swing Time of the right foot}}{\emph{Duration (s) of Stride Time of the right foot} } $ \\ \emph{(Mean of the Percentage (\%) of the Swing Time of the right foot) \cite{yogev2005dual,hausdorff1998gait}}}\\
		\\
		\hline
		\\
		$z_6$ & \makecell{Mean in duration (s) of the Swing Time \\ of the left Foot} & \makecell{\emph{Time of the left foot was in the air} \\ \emph{(Mean of the duration (s) of the Swing Time of the left foot) \cite{hausdorff2001gait,yogev2005dual,frenkel2005effect}}}\\
		\\
		\hline
		\\
		$z_7$ & \makecell{Mean in duration (s) of the Swing Time \\ of the Right Foot} & \makecell{\emph{Time of the right foot was in the air} \\ \emph{(Mean of the duration (s) of the Swing Time of the right foot) \cite{hausdorff2001gait,yogev2005dual,frenkel2005effect}}}\\
		\\
		\hline
		\\
		$z_8$ & \makecell{Coefficients of Variations of the \\ Short Swing Time} & $ 100 \times \frac{\emph{Standard Deviation of the Short Swing Time}}{\emph{Mean of the Short Swing Time}} $ \cite{yogev2007gait}\\
		\\
		\hline
		\\
		$z_9$ & \makecell{Coefficients of Variations of the \\ Long Swing Time}  & $ 100 \times \frac{\emph{Standard Deviation of the Long Swing Time}}{\emph{Mean of the Long Swing Time}} $ \cite{yogev2007gait}\\
		\\
		\hline
		\\
		$z_{10}$ & \makecell{Coefficients of Variations of the \\ Gait Asymmetry}  & $ 100 \times |ln(\frac{\emph{Coefficient of Variation of the Short Swing Time}}{\emph{Coefficient of Variation of the Long Swing Time}})| $ \cite{yogev2007gait}\\
		\\
		\hline
		\\
		$z_{11}$ & \makecell{Coefficient of Variations in duration (s) \\ of the Swing Time of the Left foot}  &
		$ 100 \times \frac{\emph{Standard Deviation in duration (s) of the Swing Time of the left foot}}
		{\emph{Mean in duration (s) of the Swing Time of the left foot}} $ \cite{yogev2007gait}\\
		\\
		\hline
		\\
		$z_{12}$ & \makecell {Coefficient of Variations in duration (s) \\ of the Swing Time of the
		Right foot} & $ 100 \times \frac{\emph{Standard Deviation in duration (s) of the Swing Time of the Right foot}}
		{\emph{Mean in duration (s) of the Swing Time of the Right foot}} $ \cite{yogev2007gait}\\
		\\
		\hline
		\\
		$z_{13}$ & \makecell {Coefficient of Variations in duration (s) \\ of the Stride Time of the
		Left foot} & $ 100 \times \frac{\emph{Standard Deviation in duration (s) of the Stride Time of the left foot}}
		{\emph{Mean in duration (s) of the Stride Time of the left foot}} $ \cite{yogev2005dual}\\
		\\
		\hline
		\\
		$z_{14}$ & \makecell {Coefficient of Variations in duration (s) \\ of the Stride Time of the
		Right foot} & $ 100 \times \frac{\emph{Standard Deviation in duration (s) of the Stride Time of the right foot}}
		{\emph{Mean in duration (s) of the Stride Time of the right foot}} $ \cite{yogev2005dual}\\
		\\
       \hline
       \\
    \end{tabular}

\end{table}

\begin{table}[t]
\scriptsize
    \caption[c]{Final clinical features based on the initial list presented in Table \ref{table_features}.}
    \centering
    \begin{tabular}{cc}
        \hline
        Feature Symbol & Feature Value  \\
        \hline
        $x_1$ & $z_1$ \\
        $x_2$ & $z_2$ \\
        $x_3$ & $z_3$ \\
        $x_4$ & $\frac{z_4+z_5}{2}$ \\
        $x_5$ & $\frac{z_6+z_7}{2}$ \\
        $x_6$ & $z_8$ \\
        $x_7$ & $z_9$ \\
        $x_8$ & $z_{10}$ \\
        $x_9$ & $\frac{z_{11}+z_{12}}{2}$ \\
        $x_{10}$ & $\frac{z_{13}+z_{14}}{2}$ \\
       \hline
    \end{tabular}
    \label{table_features2}
\end{table}

\begin{figure}[t]
\centering
\begin{tabular}{c}
    \subfigure[][]{\includegraphics[width = 4in]{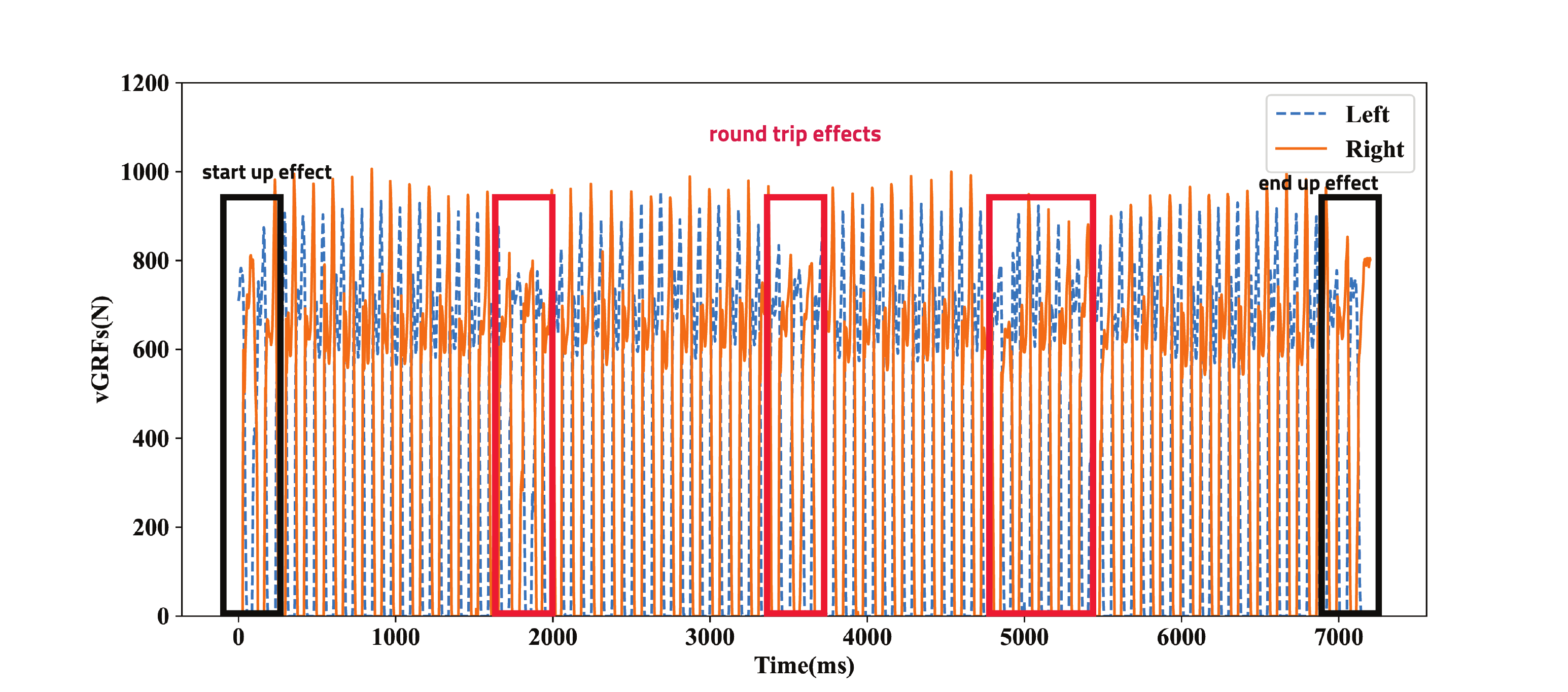}} \\
    \subfigure[][]{\includegraphics[width = 4in]{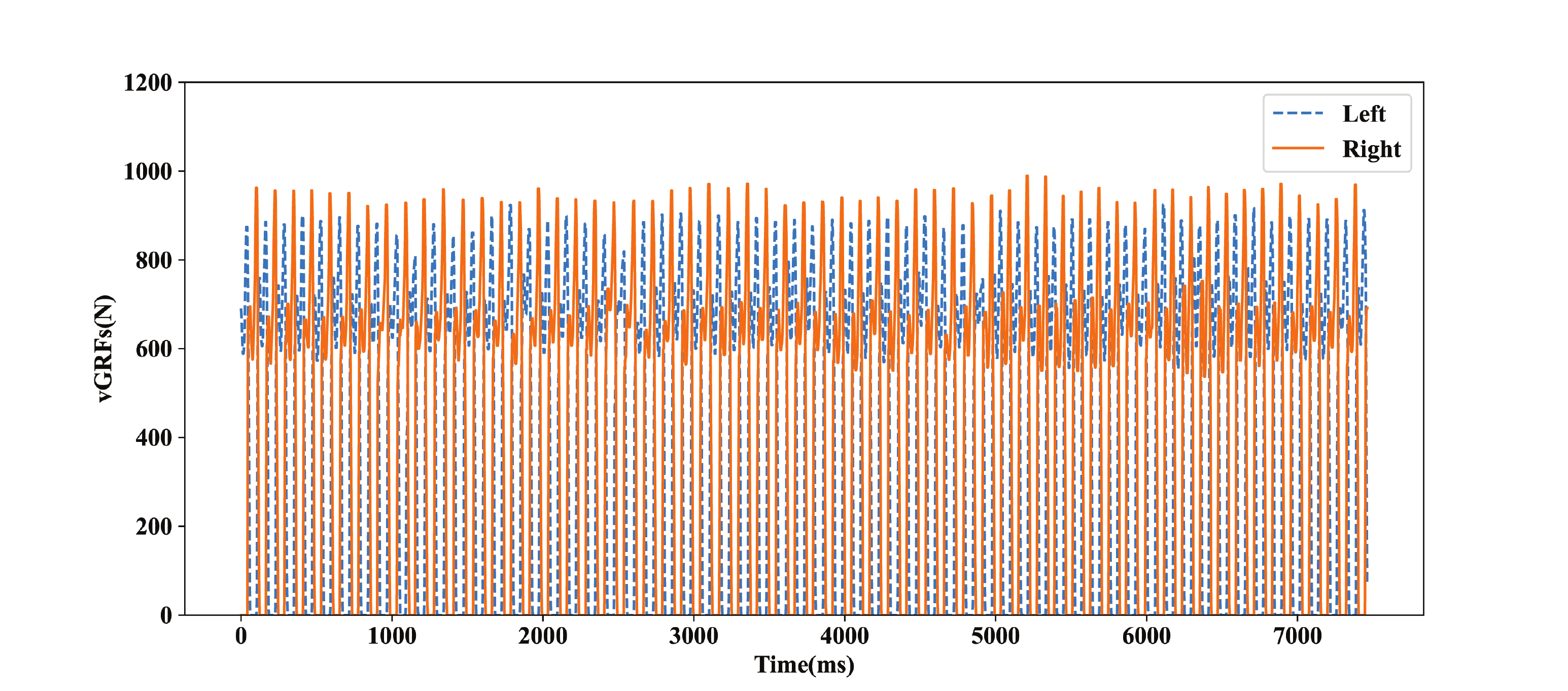}}
\end{tabular}
\caption{Effect of applying preprocessing on vGRF signals of a subject. (a) before preprocessing, (b) after preprocessing.}
\label{fig_pre}
\end{figure}

Some preprocessing is applied on the obtained vGRF data. First, following \cite{khoury2019data}, to avoid outliers resulted from performing round trip along a walkaway, the recorded gait data during the turn-around phase were removed. Moreover, first and last 20s are removed to eliminate the effects of starting and stopping movement. In addition, a 10-point median filter is applied to remove some fluctuations which can causes non-zero vGRF values during the swing phases. Figure \ref{fig_pre} shows the effect of applying the preprocessing on recorded vGRF signals of a subject.

To have a common range of definition, considering different range for different features, we normalize all features to range [0,1].

\subsection{Interval Type-2 Fuzzy Neural Network}
\label{sec43}
The general architecture of the proposed interval type-2 fuzzy neural network (IT2FNN) is shown in Figure \ref{fig_FNN}. The inputs to this network are extracted features explained in section \ref{sec41}, and its output is the degree in range [-1,1] indicating how much the subject is suspicious to have Parkinson's Disorder. This IT2FNN is built upon the Mamdani Fuzzy Inference System \cite{Mamdani}, using Minimum function as the t-norm operator. Therefore, it reduces the restriction on firing fuzzy rules. Furthermore, it is assumed that the fuzzy membership functions are Gaussian with uncertain width. Indeed, based on the proposed learning method (explained in section \ref{sec42}) the centers of fuzzy sets are extracted but there is no clue for the precise values of their width. To solve this problem, the width of fuzzy sets considered uncertain. This assumption improves the robustness of the model against uncertainty and sensor noisy measurements as investigated in section \ref{sec5}. For type reduction, the direct BMM method \cite{Biglarbegian1,Biglarbegian2,mendel2017uncertain} is applied with equal weight parameters \cite{Khanesar2016}. The details of each layer are explained as follows:

 \begin{figure}
\centering
    \includegraphics[width = 4in]{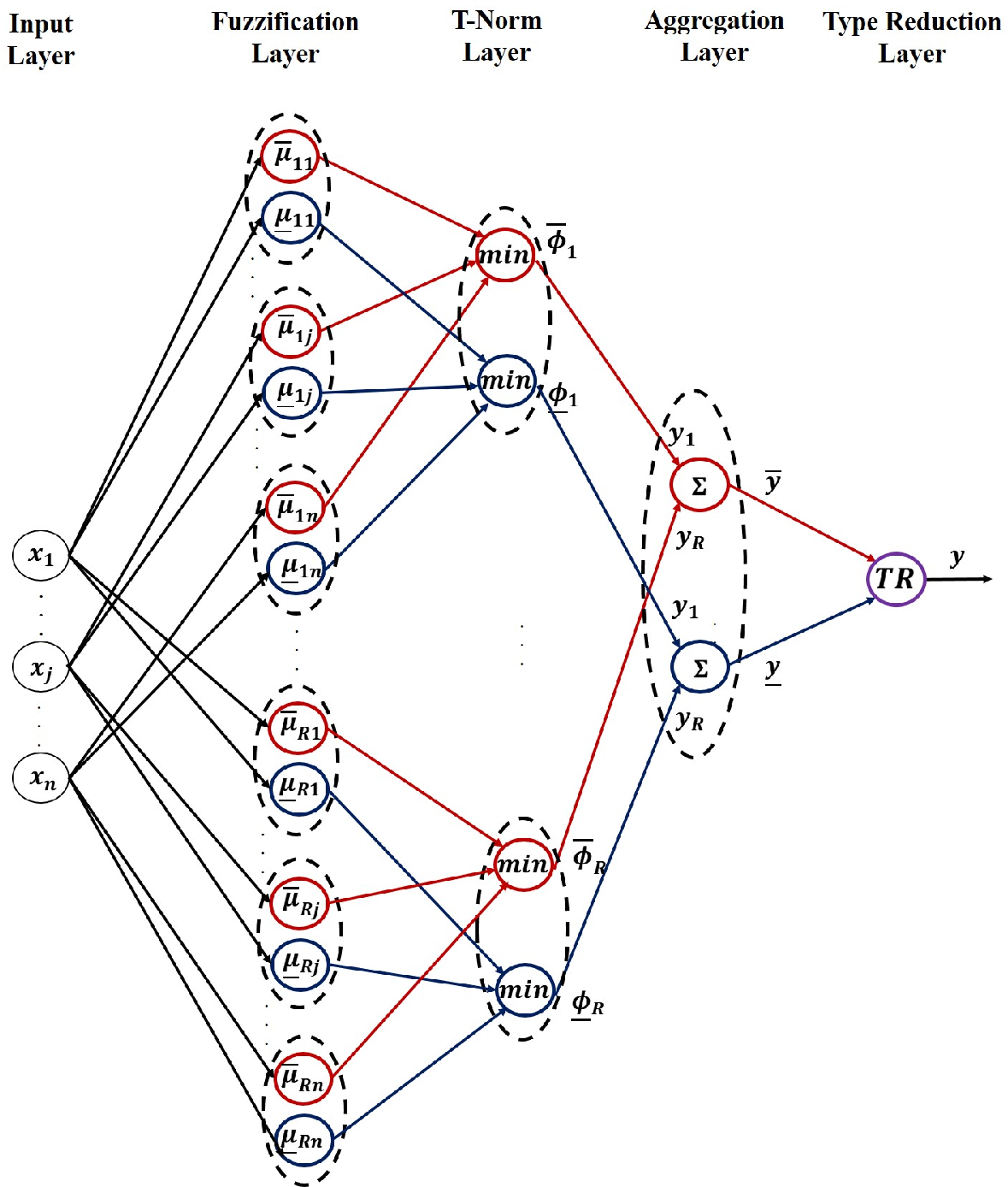}
\caption{The proposed interval type-2 fuzzy neural network architecture.}
\label{fig_FNN}
\end{figure}

\begin{enumerate}
  \item \textit{Input Layer}: This layer encodes the input variable vector extracted form the gait cycle of the subject. The output of this layer is as follows:
      \begin{equation}\label{eq1}
        X = [x_1, x_2, \cdots, x_{10}]^T
      \end{equation}
      where, $x_1$ to $x_{10}$ are extracted features from the gait cycle introduced in Table \ref{table_features2}.
  \item \textit{Fuzzification Layer}: There are two types of neurons in this layer to encode the interval type-2 fuzzy sets used for defining different fuzzy rules. For $i^{th}$ (i = 1,2,...,R) fuzzy rule, the interval type-2 fuzzy membership value of $j^{th}$ (j=1,2,...,10) input variable is defined as follows:
   \begin{equation}\label{eq2}
        \underline{\mu}_{i,j}(x_{j}) = e^{-\frac{1}{2}\frac{((x_{j}-c_{ij})^2)}{\underline{\sigma}^2}}
   \end{equation}

   \begin{equation}\label{eq3}
        \overline{\mu}_{i,j}(x_{j}) = e^{-\frac{1}{2}\frac{((x_{j}-c_{ij})^2)}{\overline{\sigma}^2}}
   \end{equation}

      where, $\underline{\mu}_{i,j}$, and $\overline{\mu}_{i,j}(x_{j})$ are lower and upper membership functions (LMF and UMF) of $j^{th}$ input variable to the $i^{th}$ fuzzy rule indicating the footprint of uncertainty (FOU), $c_{ij}$ is the center of the defined fuzzy set for $j^{th}$ input variable in the $i^{th}$ fuzzy rule, $\underline{\sigma}$ is the width of lower membership functions, and $\overline{\sigma}$ is the width of upper membership functions.
   \item \textit{T-Norm Layer}: The neurons of this layer calculate the interval of fuzzy rules' firing strength by applying minimum as the t-norm operator. Indeed, the upper and lower firing strength of each fuzzy rule is computed based on applying t-norm operator on upper and lower membership values calculated in the previous layer as follows:

        \begin{equation}
        \underline{\phi}_{i} = Min_{j=1}^{n}\underline{\mu}_{i,j}(x_{j})
        \label{eq4}
        \end{equation}

       \begin{equation}
        \overline{\phi}_{i} = Min_{j=1}^{n}\overline{\mu}_{i,j}(x_{j})
        \label{eq5}
        \end{equation}
        where $\underline{\phi}_{i}$ and $\overline{\phi}_{i}$ are lower and upper firing strength of the $i^{th}$ fuzzy rule.
      \item \textit{Aggregation Layer}: Neurons of this layer calculates the boundaries of the network's output $[\underline{y},\overline{y}]$ receiving the firing strength interval calculated in the T-Norm Layer as follows:
      \begin{equation}
        \left\{ \begin{array}{cc}
            \underline{y} &= \frac{\sum_{i=1}^{R}y_i\underline{\phi}_i}{\sum_{i=1}^{R}\underline{\phi}_i}\\
            \overline{y} &= \frac{\sum_{i=1}^{R}y_i\overline{\phi}_i}{\sum_{i=1}^{R}\overline{\phi}_i}\\
        \end{array}
        \right.
    \label{eq6}
    \end{equation}
    where, the consequent part's parameter of $i^{th}$ fuzzy rule shown by $y_i$ is a real value in the range $[-1,1]$ indicating the degree of being a patient ($y_i = 1$ indicates the patient class and $y_i= -1$ indicated the healthy one).

    \item \textit{Type Reduction Layer}: This layer is composed of one neuron that decides whether the input vector belongs to a patient or a healthy subject.  The output neuron calculated the weighted average of fuzzy rules outputs as follows:
       \begin{equation}
        y = \underline{y} + \overline{y}
    \label{eq7}
    \end{equation}
    if $y > 0$, the final decision would be patient, else the input sample is classified as healthy.
\end{enumerate}

It is worthy to mention that two final layers (aggregation and type-reduction layers) realizes the direct BMM type-reduction and defuzzification with equal weights \cite{Biglarbegian1,Biglarbegian2,Khanesar2016,mendel2017uncertain}.

\begin{figure}
\centering
    \includegraphics[width = 2.5in]{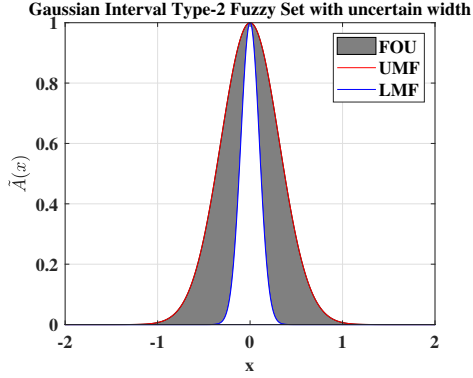}
\caption{An example of an interval type-2 fuzzy set with uncertain width. UMF: Upper Membership Function, LMF: Lower Membership Function, FOU: Footprint of Uncertainty.}
\label{fig_fs}
\end{figure}

This  Interval Type-2 Fuzzy Neural Network realizes an Interval Type-2 \cite{liang2000interval} Mamdani Fuzzy Inference Systems (FIS) \cite{Mamdani} and its fuzzy rules can be interpreted in the form of:

$$Fuzzy \, Rule \, i: IF \, x_1 \, is  \, \tilde{A}_{i,1} \, and \, x_2 \, is \,  \tilde{A}_{i,2} \, and \dots \, and \, x_{10} \, is \, \tilde{A}_{i,10} \, Then \, y \, is \, y_i$$

where, $\tilde{A}_{i,j}$ (j = 1, 2, ..., 10) is a Gaussian interval type-2 fuzzy set with uncertain width (see Figure \ref{fig_fs}) defined for $i^{th}$ fuzzy rule and $j^{th}$ variable ($x_j$ in Table \ref{table_features2}), $y$ is the output of the FIS, and $y_i \in [-1,1]$ is the consequent part of $i^{th}$ fuzzy rule. To report the fuzzy rules, it is sufficient to report the extracted centers for all interval type-2 fuzzy sets $\tilde{A}_{i,j}$.

\subsection{Learning Algorithm}
\label{sec42}
The introduced network in section \ref{sec43} has two categories of parameters we should determine to classify subjects properly: 1- Fuzzy rules antecedent parts' parameters including the centers and width of fuzzy sets, 2- Fuzzy rules consequent parts' parameters indicating the decision of each fuzzy rule ($y_i$). We determine these parameters by applying a centroid-based clustering method, like Fuzzy C-Means \cite{FCM} on training samples as follows:

\begin{enumerate}
  \item Divide the training samples into $R$ clusters. Here, the desired output (label) which is $+1$ for patients and $-1$ for healthy subjects is considered as an input variable. Therefore, the $k^{th}$ training sample $X_k$ has is a vector with 11 dimensions as follows:
      \begin{equation}\label{eq8}
        X_k = [x_{k,1}, x_{k,2}, \dots, x_{k,10}, y_{k}^*]
      \end{equation}
      where $x_{k,j}$ (j = 1,2, ..., 10) are values of input variables for $k^{th}$ sample and $y_k^*$ is its desired output value. Using the output value along with input variables causes that the clustering approach considers the classes of training instances in composing clusters.
  \item After clustering, $i^{th}$ (i = 1,2, ..., R) cluster has a center vector $V_i$ representing that cluster. These centers are used as the centers of fuzzy sets composing each fuzzy rule.
  \item To determine the consequent part parameters $y_i$ for $i^{th}$ fuzzy rule, a weighted average of desired outputs of training samples belongs to the $i^{th}$ cluster is calculated as follows:
            \begin{equation}\label{eq8}
        y_i = \frac{\sum_{k=1}^{N}u_{i,k}^m.y_k}{\sum_{k=1}^{N}u_{i,k}^m}
      \end{equation}
      where $y_k^*$ is desired output value for $k^{th}$ training sample, $u_{i,k}$ is the fuzzy membership value of $k^{th}$ training sample to the $i^{th}$ cluster calculated during the clustering, $N$ is the number of training samples, and $m>1$ is the fuzzy partition exponent for controlling the degree of fuzzy overlap in FCM algorithm.
  \item To determine the fuzzy sets width, considering the uncertainty we assume that the width belongs to the range $[\sigma_1, \, \sigma_2]$ where $\sigma_1$ and $\sigma_2$ are two constant values lower than 1. Therefore, $\sigma_1$ is considered as the width of the lower membership functions, and $\sigma_2$ is considered as the width of the upper membership function as follows:
      \begin{equation}\label{eq9}
      \begin{array}{c}
        \underline{\sigma} = \sigma_1\\
       \overline{\sigma} = \sigma_2
      \end{array}
      \end{equation}
\end{enumerate}

The proposed learning algorithm is summarized in Algorithm \ref{alg1}.

\begin{algorithm}[t]
\scriptsize
  \textbf{Inputs:} \\
  $\quad$ \text{training samples $X$ along with their desired output values $Y^*$} \\
  $\quad$ \text{number of rules ($R$), number of samples ($N$)} \\
  $\quad$ \text{range of fuzzy sets width ($\sigma_1$,$\sigma_2$), fuzzy overlap degree m} \\
  \textbf{Outputs:}\\
   $\quad$ \text{center of fuzzy sets ($c_{i,j}$), width of lower and upper membership functions ($\underline\sigma$ $\&$ $\overline\sigma$)}\\
   $\quad$ \text{the consequent parts' parameters $y_i$}\\
  $X_{aug} \gets [X,Y^*]$ \;
  \text{Calculates the cluster centers $V$ and membership matrix $u$ based on FCM:} \\
  $\quad$ $[V,u] \gets FCM(X_{aug},m,R)$\;

    \For {$i \gets 1$ \KwTo $R$}{
    \For {$j \gets 1$ \KwTo $10$}{
        $c_{i,j} \gets V_{i,j}$ \;
    }
    $y_i \gets 0$ \;
    $s \gets 0$ \;
    \For {$k \gets 1$ \KwTo $N$}{
        $y_i \gets y_i + u_{i,k}^m*Y^*_k$ \;
        $s \gets s + u_{i,k}^m$ \;
    }
    $y_i \gets y_i/s$ \;
    }

    $\underline{\sigma} \gets \sigma_1 $\;
        $\overline{\sigma} \gets \sigma_2 $\;
 \caption{Learning algorithm}
 \label{alg1}
 \end{algorithm}

\subsection{Complementary Online Learning}

\begin{figure}[t]
\centering
    \includegraphics[width = 4.5in]{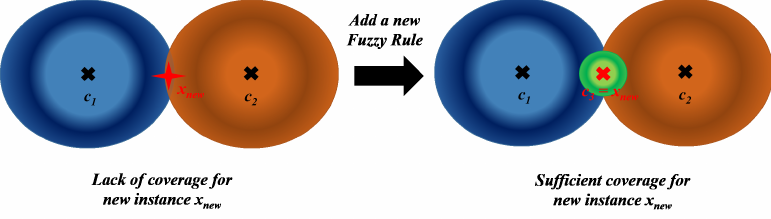}
\caption{An illustrative example of adding a new fuzzy rule in a two-dimensional input space. First, the input vector $x_{new}$is analyzed by two existing fuzzy rules with centers $c_1$ and $c_2$. The total coverage for the sample $x_{new}$ is lower than a predefined threshold $\theta_c$. Consequently, a new fuzzy rule, which its center is $x_{new}$, has been added.}
\label{fig_add}
\end{figure}

\begin{figure}[t]
\centering
    \includegraphics[width = 4.5in]{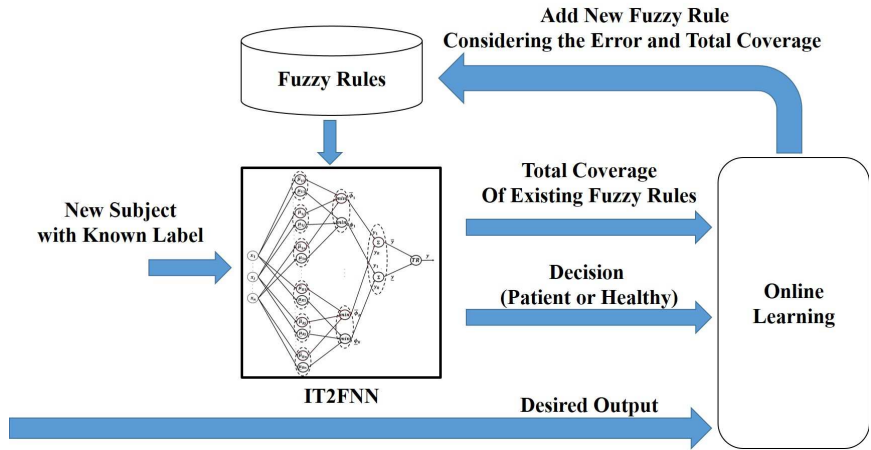}
\caption{The paradigm of online learning.}
\label{fig_online}
\end{figure}

The proposed learning method is a kind of batch learning approach based on clustering all samples which assumes that all training instances are available at once.
This is partially true that we should have training examples for training the fuzzy neural network at first. However, it is possible that more studies may gather more examples in the future. Therefore, to avoid learning the network from scratch, here we propose a complementary learning approach to able the network handing the new training instances whenever is necessary.
The proposed complementary online learning approach is based on investigating the amount of current fuzzy rules' coverage encountering a new instance \cite{Salimi2020novel,Salimi-Badr2017}. If the coverage of existing fuzzy rules is not sufficiently encountering a new example, a new fuzzy rule is added to the Fuzzy Neural Network to cover an area consisting of this new sample and future similar instances.
Indeed, if the sum of fuzzy rules firing strength is lower than a predefined threshold, a new fuzzy rule based on this new sample is added to the network. The center of this newly added fuzzy rule would be the arrived new instance and the width would be lower than the previous fuzzy rules to decrease the conflict with previous fuzzy rules and only cover the region with a low amount of coverage. The consequent part of this new fuzzy rule is the desired output of the newly arrived instance. Figure \ref{fig_add} shows an illustrative example of adding a new fuzzy rule in a two-dimensional space and Algorithm \ref{alg2} summarizes this complementary online learning. The paradigm of complementary online learning is shown in Figure \ref{fig_online}.

\begin{algorithm}[t]
\scriptsize
\textbf{Inputs:} \\
  $\quad$ \text{New training sample $x_{new}$ along with its desired output values $y_{new}^*$} \\
  $\quad$ \text{number, centers, width and consequent parameters of existing rules ($R,c,\overline{\sigma},\underline{\sigma},y$)} \\
  $\quad$ \text{threshold of coverage $\theta_c$} \\
  $\quad$ \text{coefficient $\epsilon$ to justify the width of added fuzzy rules} \\
  \textbf{Outputs:}\\
   $\quad$ \text{updated center of fuzzy sets ($c_{i,j}$), width of lower and upper membership functions ($\underline{\sigma}$ $\&$ $\overline{\sigma}$)}\\
	$\quad$ \text{updated consequent parts' parameters $y_i$} \\
\text{compute the decision of fuzzy neural network for $x_{new}$ as $\hat{y}_{new}$} \\

\If {$\hat{y}_{new} \neq y^*_{new}$}{
$S \gets 0$\;
\For {$i \gets 1$ \KwTo $R$}{
$S \gets S + \phi_i(x_{new})$ \;
}

\If{$S < \theta_c$}{
		$R \gets R + 1$\;
		\For{$j \gets 1 \quad$ \KwTo $\quad n$}{
		$c_{R,j} \gets x_{new,j}$ \;
		}
		$y_{R} \gets y^*_{new}$\;
		$\overline{\sigma}_{R} \gets \epsilon \times \overline{\sigma}$\;
		$\underline{\sigma}_{R} \gets \epsilon \times \underline{\sigma}$\;
		}
}
\caption{Complementary Online Learning Algorithm}
 \label{alg2}
\end{algorithm}

\section{Experimental Results}
\label{sec5}
In this section, the performance of the proposed method is evaluated and compared to some previous methods. First, the utilized data is explained. Next, the evaluation metrics used for evaluating the performance of the method are introduced. Afterward, the performance of the proposed paradigm is compared to some supervised and unsupervised machine learning approaches applied to similar data in the literature. Next, the effectiveness of the proposed complementary online learning method is investigated. To investigate the effectiveness of using Interval Type-2 Fuzzy Logic, the performance of the proposed method is compared with the performance of a similar type-1 fuzzy neural network encountering noisy data. Afterward, the final fuzzy sets and rules extracted by the proposed method are reported. These reported fuzzy rules can be used by experts to: 1- evaluate the extracted fuzzy rules and fine-tune them, and 2- apply them to classify patients and healthy persons.

\subsection{Data}
We utilized the gait cycle data provided by PhysioNet\footnote{https://physionet.org/content/gaitpdb/1.0.0/}. This data aggregates three similar sub-datasets obtained in different experimental studies from different subjects \cite{frenkel2005treadmill,yogev2005dual,hausdorff2007rhythmic}. The dataset includes vGRF obtained from 16 sensors placed under feet of control and patient subjects (8 per foot) as shown in Fig. \ref{fig1}.  Totally, the data were recorded from 93 patients with PD and 72 healthy control subjects. The vertical ground reaction force recorded of subjects as they walked at their usual, self-selected pace for approximately 2 minutes on level ground. Features are extracted from two sequences indicating the average vGRF of each foot. The details of each dataset are outlined in Table \ref{table_data}.

\begin{table}[t]
\scriptsize
    \caption[c]{Information of different datasets used for evaluation of the method (CO: Healthy Control Subject, PD: Patients with Parkinson's Disease).}
    \centering
    \begin{tabular}{ccccc}
        \hline
        Data & Subjects & Total subjects & Female & Male \\
        \hline
        Ga \cite{yogev2005dual} & PD & 29 & 9 & 20\\
        & CO & 18 & 8 & 10\\
        Si \cite{frenkel2005treadmill} & PD & 35 & 13 & 22\\
        & CO & 29 & 11 & 18\\
        Ju \cite{hausdorff2007rhythmic} & PD & 29 & 13 & 16\\
		& CO & 25 & 14 & 12\\
       \hline
    \end{tabular}
    \label{table_data}
\end{table}

\subsection{Evaluation Metrics}
To evaluate the performance of the proposed method, the usual metrics utilized in the classification tasks are used. One of the most popular metrics in classification tasks is "Accuracy" that shows the average performance of the method to properly classify instances and is defined as follows:

\begin{equation}
    Accuracy = \frac{TP+TF}{TP+TN+FP+FN}
\end{equation}
where TP ("True Positive") is the number of instances that belongs to the positive class (here PD patient) classified correctly, TN (True Negative) is the number of instances belongs to the negative class (here healthy persons) classified correctly, FP (False Positive) is the number of instances belongs to the negative class but classified wrongly, and finally, FN (False Negative) is the number of instances belonging to the positive class but classified wrongly.

In some applications like medical diagnosis, it is very important to avoid False Positives and also detect the patients correctly. Detecting a healthy person as a patient would lead to extra costs (e.g. additional experiments) and also disappointment for the person and his/her family. On the other hand, if the system detects a patient as a healthy person, it would be dangerous for his/her health. Therefore, to have a more precise evaluation, the following metrics are also utilized:

\begin{align}
    Precision = \frac{TP}{TP+FP} \\
    Recall = \frac{TP}{TP+FN} \\
    F_1 = \frac{2.Precision.Recall}{Precision+Recall}
\end{align}
where, Precision evaluates the False Positive rate, Recall evaluates the ability of the method to classify patients correctly, and $F_1$ score is a combination of Precision and Recall.

\subsection{Results}
Tables \ref{table_res1} to \ref{table_res3} evaluate the performance of the method based on different criteria in different datasets introduced in Table \ref{table_data}. The performance of the proposed method is compared with the performance of the methods used similar clinical features in \cite{khoury2019data}. Following \cite{khoury2019data}, each sub-dataset is divided into training and testing sets according to the leave-one-out cross-validation procedure.

\begin{table}[t]
\scriptsize
    \caption[c]{Comparison of the performance of the proposed method with other approaches in the case of the Yogev et al. (Ga) sub-datasets.}
    \centering
    \begin{tabular}{ccccccccc}
        \hline
        Performance Metric & Propose Method & KNN \cite{khoury2019data} &  CART \cite{khoury2019data} & RF \cite{khoury2019data}  &NB \cite{khoury2019data} & SVM \cite{khoury2019data} & K-Means \cite{khoury2019data} &GMM \cite{khoury2019data}\\
        \hline
        Accuracy & 92.92 & 86.05 & 83.72 & 86.05 & 74.42 & 86.05 & 63.72 & 64.77\\
		Precision & 92.40 & 84.89 & 82.86 & 85.07 & 74.73 & 84.90 & 64.34 & 62.63\\
		Recall & 97.33 & 86.34 & 81.94 & 85.07 & 76.45 & 85.71 & 65.31 & 62.91\\
		F1 Score & 94.80 & 85.61 & 82.40 & 85.07 & 75.58 & 85.30 & 64.82 & 62.77\\
       \hline
    \end{tabular}
    \label{table_res1}
\end{table}

\begin{table}[t]
\scriptsize
    \caption[c]{Comparison of the performance of the proposed method with other approaches in the case of the Hausdordd et al. (Ju) sub-datasets.}
    \centering
    \begin{tabular}{ccccccccc}
        \hline
        Performance Metric & Propose Method & KNN \cite{khoury2019data} &  CART \cite{khoury2019data} & RF \cite{khoury2019data}  &NB \cite{khoury2019data} & SVM \cite{khoury2019data} & K-Means \cite{khoury2019data} &GMM \cite{khoury2019data}\\
        \hline
        Accuracy & 85.83 & 90.91 & 84.30 & 87.60 & 77.69 & 90.08 & 55.12 & 65.12\\
		Precision & 88.89 & 85.35 & 82.34 & 89.41 & 70.64 & 89.31 & 52.58 & 57.95\\
		Recall & 93.62 & 88.35 & 64.96 & 71.48 & 78.54 & 78.96 & 53.91 & 61.29\\
		F1 Score & 91.19 & 86.83 & 72.62 & 79.45 & 74.38 & 83.82 & 53.24 & 59.57\\
       \hline
    \end{tabular}
    \label{table_res2}
\end{table}

\begin{table}[t]
\scriptsize
    \caption[c]{Comparison of the performance of the proposed method with other approaches in the case of the Frenkel-Toledo et al. (Si) sub-datasets.}
    \centering
    \begin{tabular}{ccccccccc}
        \hline
        Performance Metric & Propose Method & KNN \cite{khoury2019data} &  CART \cite{khoury2019data} & RF \cite{khoury2019data}  &NB \cite{khoury2019data} & SVM \cite{khoury2019data} & K-Means \cite{khoury2019data} &GMM \cite{khoury2019data}\\
        \hline
        Accuracy & 83.33 & 81.25 & 79.69 & 82.81 & 79.69 & 82.81 & 57.19 & 65.31\\
		Precision & 85.71 & 81.43 & 79.56 & 83.10 & 79.69 & 82.81 & 61.07 & 64.95\\
		Recall & 85.71 & 81.67 & 79.36 & 82.22 & 79.95 & 83.10 & 59.23 & 64.62\\
		F1 Score & 85.71 & 81.55 & 79.46 & 82.65 & 79.82 & 82.96 & 60.14 & 64.78\\
       \hline
    \end{tabular}
    \label{table_res3}
\end{table}

According to the results reported in Tables \ref{table_res1} to \ref{table_res3}. the proposed method outperforms all other supervised and supervised machine learning approaches trained with similar features in sub-dataset Ga, and Si. In sub-dataset Ju, two supervised machine learning approaches KNN and SVM have better Accuracy, the proposed method has a better Precision, Recall, and F1 Score. Indeed, the proposed method has a lower False Negative rate which is very important in a clinical application. Moreover, the proposed method is an interpretable approach, and experts can use the extracted fuzzy rules or fine-tune them.
Tables \ref{table_tp1} to \ref{table_tp3} show the true positive and true negative rates of the proposed method and compare its performance with other methods. The proposed method performs worse than KNN in Ju sub-dataset in detecting healthy control subjects. The reason could be low coverage of healthy input space by the gathered data in Ju sub-dataset. Therefore, there are not sufficient healthy instances to cover the input space efficiently. Subsequently, the KNN approach that saves all instances and votes in a neighborhood around the test example performs better than the proposed method in this sub-dataset.

\begin{table}[t]
\tiny
    \caption[c]{Global confusion matrix obtained using the different classifiers in the case of the Yogev et al. (Ga) sub-dataset (H: Healthy, P: Patient).}
    \centering
    \begin{tabular}{cccccccccccccccccc}
        \hline
       &&\multicolumn{16}{c}{\textbf{Obtained Classes}}\\
        \hline
		&&\multicolumn{2}{c}{\textbf{Proposed Method}}&\multicolumn{2}{c}{\textbf{KNN}}&\multicolumn{2}{c}{\textbf{CART}}&\multicolumn{2}{c}{\textbf{RF}}&\multicolumn{2}{c}{\textbf{NB}}&\multicolumn{2}{c}{\textbf{SVM}}&\multicolumn{2}{c}{\textbf{K-Means}}&\multicolumn{2}{c}{\textbf{GMM}}\\
        & & H & P & H & P & H & P & H & P & H&P&H&P&H&P&H&P\\
        \hline
        True & H & 84.21 & 15.79 & 87.50 & 12.50 & 75.00 & 25.00 & 81.25 & 18.75 & 84.38 & 15.62 & 84.38 & 15.62 & 71.53 & 28.47 & 55.63 & 44.37\\
        Classes & P & 2.67  & 97.33 & 14.81 & 85.19 & 11.11 & 88.89 &11.11 & 88.89& 31.48 & 68.52 & 12.96 & 87.04 & 40.91 & 59.09 & 29.81 & 70.19 \\
         \hline
    \end{tabular}
    \label{table_tp1}
\end{table}

\begin{table}[t]
\tiny
    \caption[c]{Global confusion matrix obtained using the different classifiers in the case of the Hausdorff et al. (Ju) sub-dataset (H: Healthy, P: Patient).}
    \centering
    \begin{tabular}{cccccccccccccccccc}
        \hline
         &&\multicolumn{16}{c}{\textbf{Obtained Classes}}\\
         \hline
         &&\multicolumn{2}{c}{\textbf{Method}}&\multicolumn{2}{c}{\textbf{KNN}}&\multicolumn{2}{c}{\textbf{CART}}&\multicolumn{2}{c}{\textbf{RF}}&\multicolumn{2}{c}{\textbf{NB}}&\multicolumn{2}{c}{\textbf{SVM}}&\multicolumn{2}{c}{\textbf{K-Means}}&\multicolumn{2}{c}{\textbf{GMM}}\\
        & & H & P & H & P & H & P & H & P & H&P&H&P&H&P&H&P\\
        \hline
        True & H & 57.69 & 42.31 & 84.00 & 16.00 & 32.00 & 68.00 & 44.00 & 56.00 & 80.00 & 20.00 & 60.00 & 40.00 & 51.84 & 48.16 & 54.76 & 45.24\\
        Classes & P & 6.38  & 93.62 & 7.29 & 92.71 & 2.08 & 97.92 & 1.04 & 98.96& 22.92 & 77.08 & 2.08 & 97.92 & 44.02 & 55.98 & 32.19 & 67.81 \\
         \hline
    \end{tabular}
    \label{table_tp2}
\end{table}

\begin{table}[t]
\tiny
    \caption[c]{Global confusion matrix obtained using the different classifiers in the case of the Frenkel-Toledo et al. (Si) sub-dataset (H: Healthy, P: Patient).}
    \centering
    \begin{tabular}{cccccccccccccccccc}
        \hline
       &&\multicolumn{16}{c}{\textbf{Obtained Classes}}\\
        \hline
		&&\multicolumn{2}{c}{\textbf{Proposed Method}}&\multicolumn{2}{c}{\textbf{KNN}}&\multicolumn{2}{c}{\textbf{CART}}&\multicolumn{2}{c}{\textbf{RF}}&\multicolumn{2}{c}{\textbf{NB}}&\multicolumn{2}{c}{\textbf{SVM}}&\multicolumn{2}{c}{\textbf{K-Means}}&\multicolumn{2}{c}{\textbf{GMM}}\\
        & & H & P & H & P & H & P & H & P & H&P&H&P&H&P&H&P\\
        \hline
        True & H & 80.00 & 20.00 & 86.21 & 13.79 & 75.86 & 24.14 & 75.86 & 24.14 & 82.76 & 17.24 & 86.21 & 13.79 & 80.97 & 19.03 & 57.24 & 42.76 \\
        Classes & P & 14.29  & 85.71 & 22.86 & 77.14 & 17.14 & 82.86 & 11.43 & 88.57& 22.86 & 77.14 & 20.00 & 80.00 & 62.51 & 37.49& 28.00 & 72.00 \\
         \hline
    \end{tabular}
    \label{table_tp3}
\end{table}

\begin{figure}[!b]
\centering
\begin{tabular}{ccc}
    \subfigure[][]{\includegraphics[width = 2.0in]{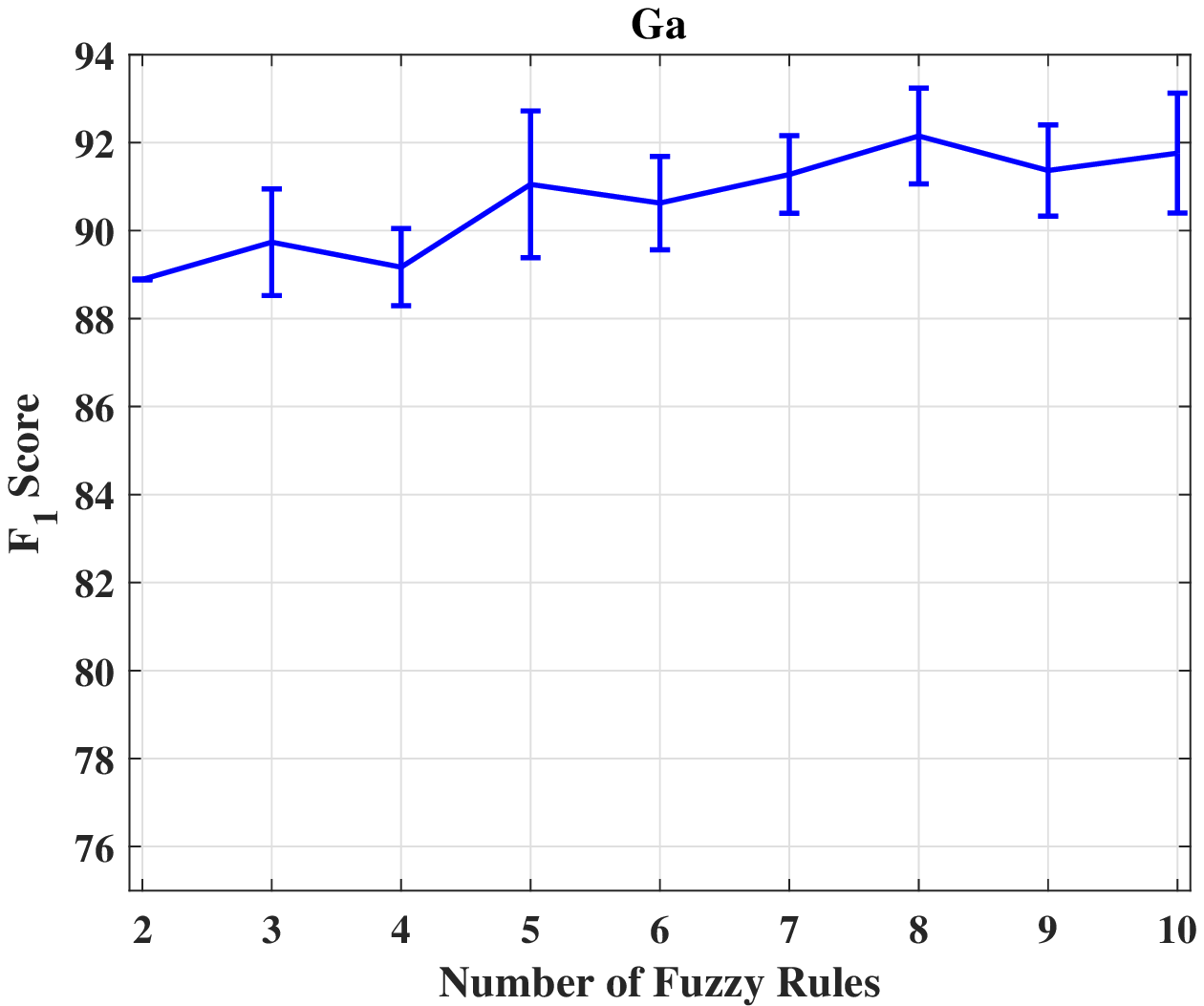}} &
    \subfigure[][]{\includegraphics[width = 2.0in]{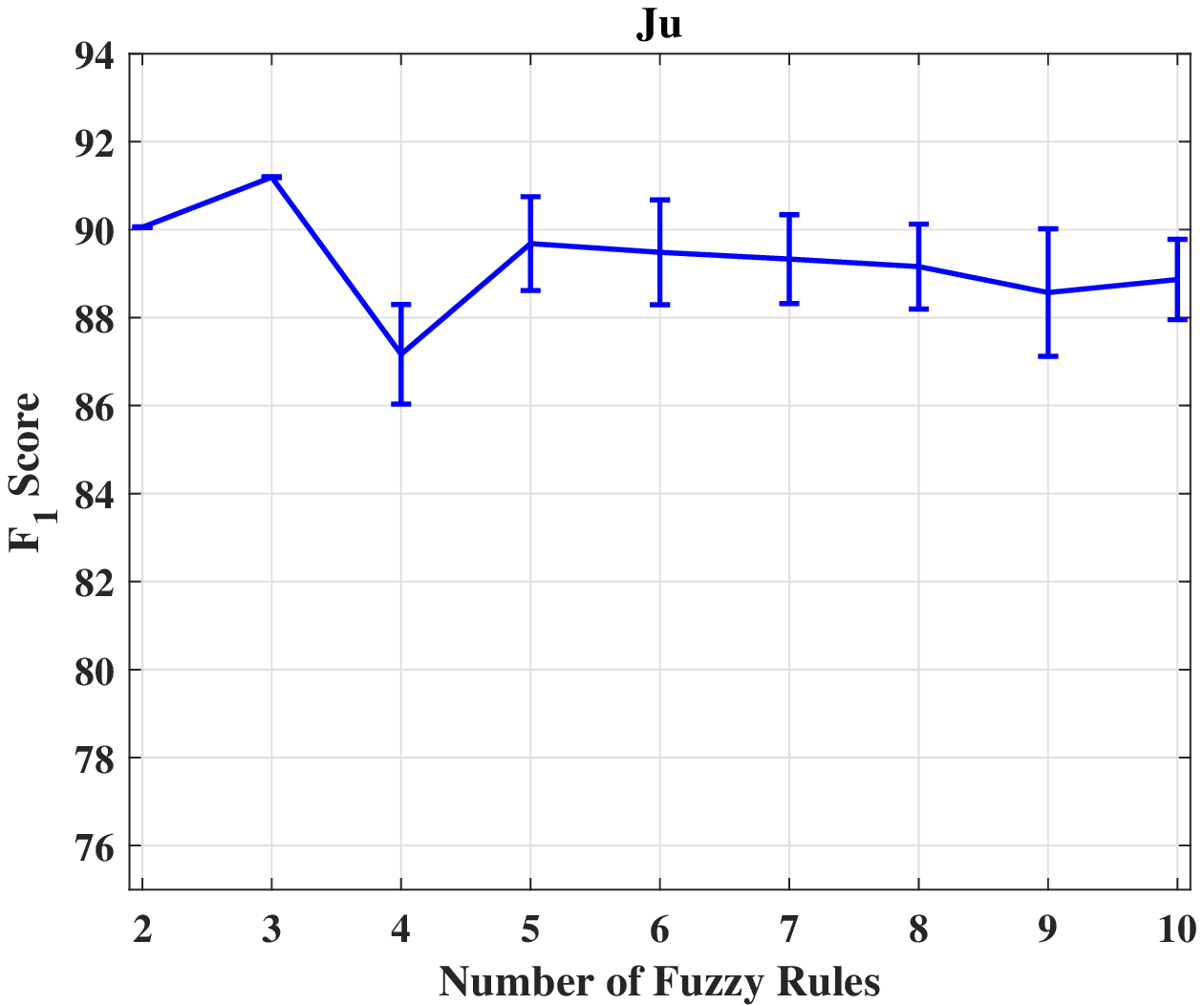}} &
    \subfigure[][]{\includegraphics[width = 2.0in]{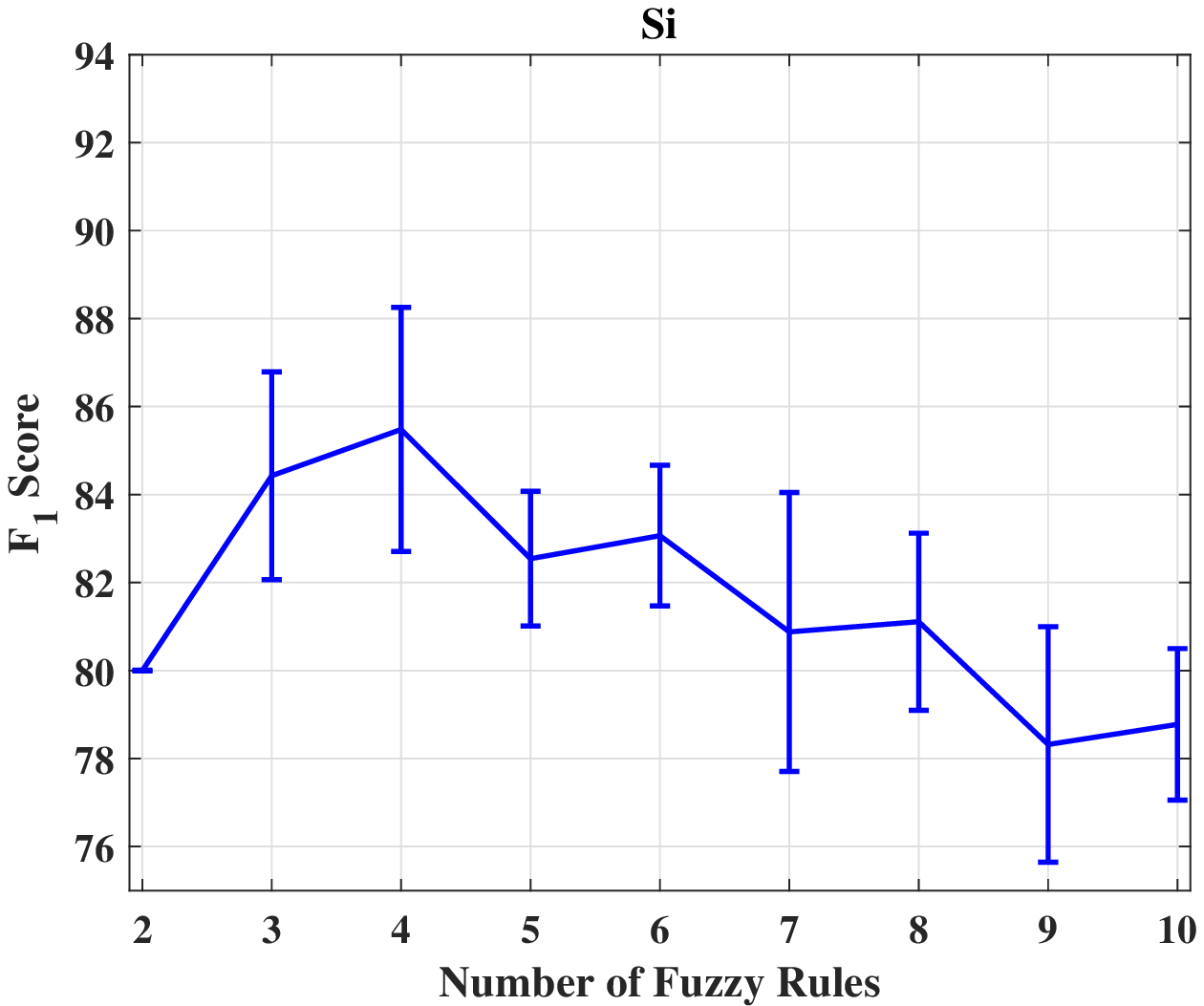}}
\end{tabular}
\caption{Mean and standard deviation of the performance of the method regarding different number of fuzzy rules.}
\label{fig_num}
\end{figure}

For parameter setting, we should determine the number of fuzzy rules and the width of fuzzy sets. The width of fuzzy sets are considered in the range $[0.01,0.1]$. Indeed, it is assumed that each fuzzy set could be very uncertain ($\sigma = 0.1$) or close to a precise value ($\sigma = 0.01$) for decision making. To determine the number of fuzzy rules, the performance of the method is evaluated for a different number of fuzzy rules for 10 independent runs and then we have calculated the mean and the standard deviation (STD) of the F1 score regarding different number of fuzzy rules. Afterward, the lowest number of fuzzy rules with higher mean and lower STD is chosen for each dataset (see Figure \ref{fig_num}). Based on Figure \ref{fig_num}, the chosen number of fuzzy rules are listed in Table \ref{table_num_rule}.

\begin{table}[t]
\scriptsize
    \caption[c]{Chosen number of fuzzy rules based on studies reported in Figure \ref{fig_num}.}
    \centering
    \begin{tabular}{cccc}
        \hline
        Data & Ga & Ju & Si \\
        \hline
        Number of Fuzzy Rules & 8 & 3 & 4 \\
       \hline
    \end{tabular}
    \label{table_num_rule}
\end{table}

\subsection{Effectiveness of the Online Learning}
To evaluate the effectiveness of the proposed complementary online learning, we assume that only one sub-set is available as the initial training dataset. After learning the Interval Type-2 Fuzzy Neural Network and extracting the fuzzy rules, another sub-dataset is added as new training samples. We assume that the sub-dataset Ga \cite{yogev2005dual} is available and after training, the samples of sub-dataset Ju will be added. Table \ref{table_online} compares the performance of the method before and after adding new instances. It is shown that the number of fuzzy rules is increased and the performance of the method is improved for all sub-datasets. The threshold for investigating the amount of fuzzy rules' coverage is set 0.1. The $\epsilon$ parameter in Algorithm \ref{alg2} is set at 1.

\begin{table}[t]
\scriptsize
    \caption[c]{Evaluating the effectiveness of online learning. The initial available training dataset includes Ga and after training, Ju is added}
    \centering
    \begin{tabular}{cc|ccccc}
        \hline
         State & Number of Fuzzy Rules &Data & Accuracy & Precision &  Recall & F1 Score\\
        \hline
         &  & Ga & 92.92 & 94.66 & 94.66 & 94.66\\
        Trained using Ga &8& Ju & 75.00 & 77.78 & 80.00 & 78.87\\
         && Si & 81.66 & 90.00 & 86.17 & 88.04\\
         \hline
         & & Ga & 93.80 & 93.59 & 97.33 & 95.42\\
         Added Ju&10& Ju & 83.33 & 80.48 & 94.28 & 86.84\\
         && Si & 87.50 & 88.34 & 96.81 & 92.38\\
       \hline
    \end{tabular}
    \label{table_online}
\end{table}

\subsection{Effectiveness of Interval Type-2 Fuzzy System against Noisy Data}
\label{noise}

\begin{table}[h]
\scriptsize
    \caption[c]{Evaluating the effectiveness of Interval Type-2 Fuzzy Neural Network against noisy data.}
    \centering
    \begin{tabular}{cccccccc}
        \hline
         Noise &  Number of Fuzzy Rules & Data & Method &Accuracy & Precision &  Recall & F1 Score\\
        \hline
         0.1  & 10 & Ga & IT2 & 91.96 & 93.42 & 94.66 & 94.04\\
          & & & T1 & 89.28 & 89.87 & 94.66 & 92.20\\
          & & Ju & IT2 & 89.74 & 91.83 & 95.74 & 93.75\\
          & & & T1 & 86.32 & 88.23 & 95.74 & 91.83\\
          & & Si & IT2 & 81.25 & 78.04 & 91.42 & 84.21\\
          & & & T1 & 81.25 & 76.74 & 94.28 & 84.61\\
         \hline
         0.3  & 10 & Ga & IT2 & 83.03 & 87.83 & 86.66 & 87.24\\
          & & & T1 & 84.82 & 88.15 & 89.33 & 88.74\\
          & & Ju & IT2 & 83.76 & 92.13 & 87.23 & 89.61\\
          & & & T1 & 80.34 & 87.36 & 88.29 & 87.83\\
          & & Si & IT2 & 76.56 & 79.41 & 77.14 & 78.26\\
          & & & T1 & 67.18 & 69.44 & 71.42 & 70.42\\
       \hline
    \end{tabular}
    \label{table_noise}
\end{table}

To investigate the effectiveness of the proposed Interval Type-2 Fuzzy system against noisy data, its performance is compared with a type-1 fuzzy neural network applying on noisy test samples. In this experiment, all samples of sub-datasets Ga, Ju, and Si are gathered together to form one dataset. Next, it is divided into training and testing sets according to the leave-one-out cross-validation procedure. We add a Gaussian noise with different width values $0.1$ and $0.3$ to the test samples. The number of fuzzy rules is considered as 10. Table \ref{table_noise} shows the effectiveness of the proposed Interval Type-2 Fuzzy Neural Network comparing its performance with a Type-1 Fuzzy Neural Network.

\begin{table}[h]
\scriptsize
    \caption[c]{Final performance of the proposed method trained using all available data}
    \centering
    \begin{tabular}{cccc}
        \hline
         Accuracy & Precision &Recall & F1 Score\\
         88.74 & 89.41 & 95.10 & 92.16\\
        \hline
    \end{tabular}
    \label{table_final}
\end{table}

\begin{figure}[!b]
\centering
\begin{tabular}{cc}
    \subfigure[][]{\includegraphics[width = 1.8in]{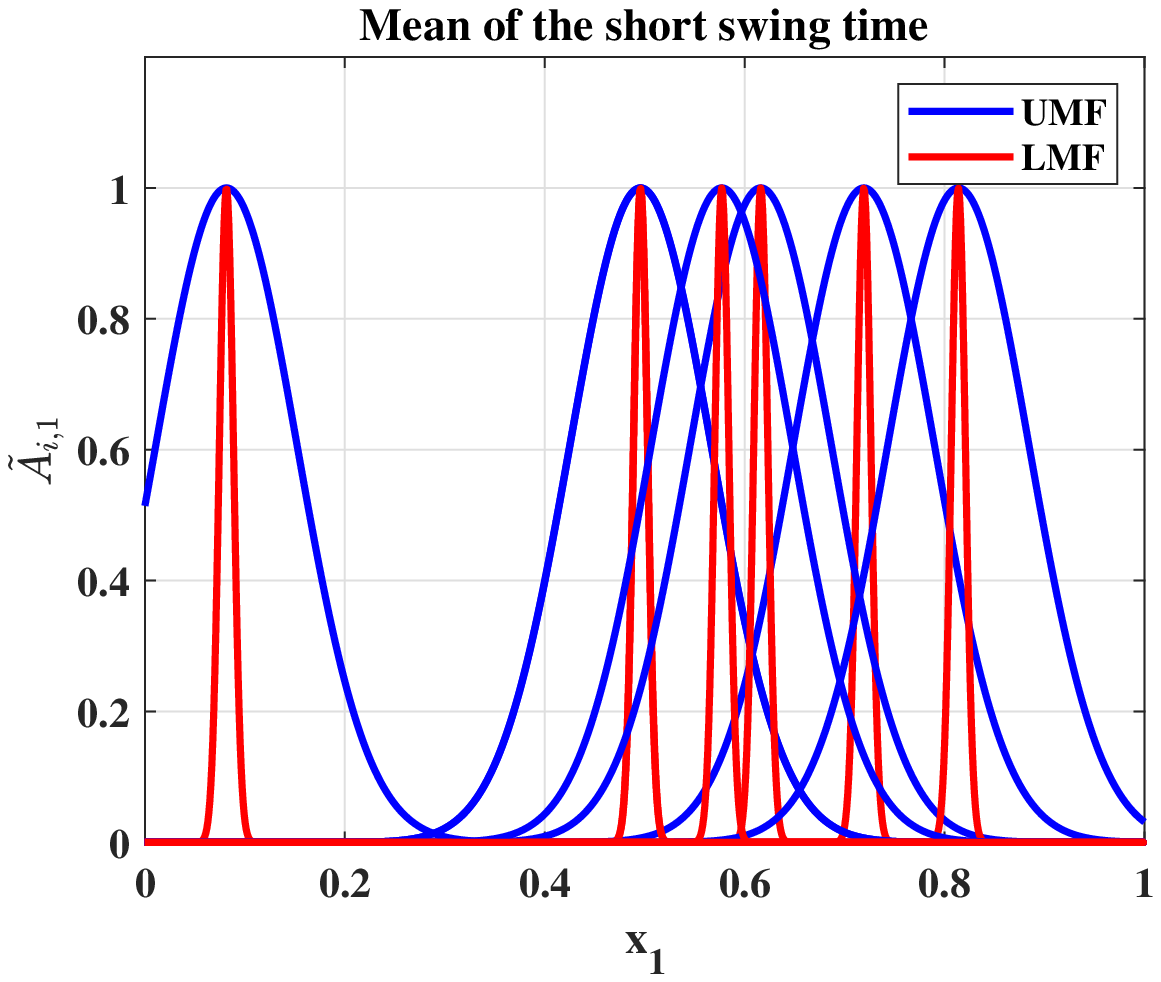}} &
    \subfigure[][]{\includegraphics[width = 1.8in]{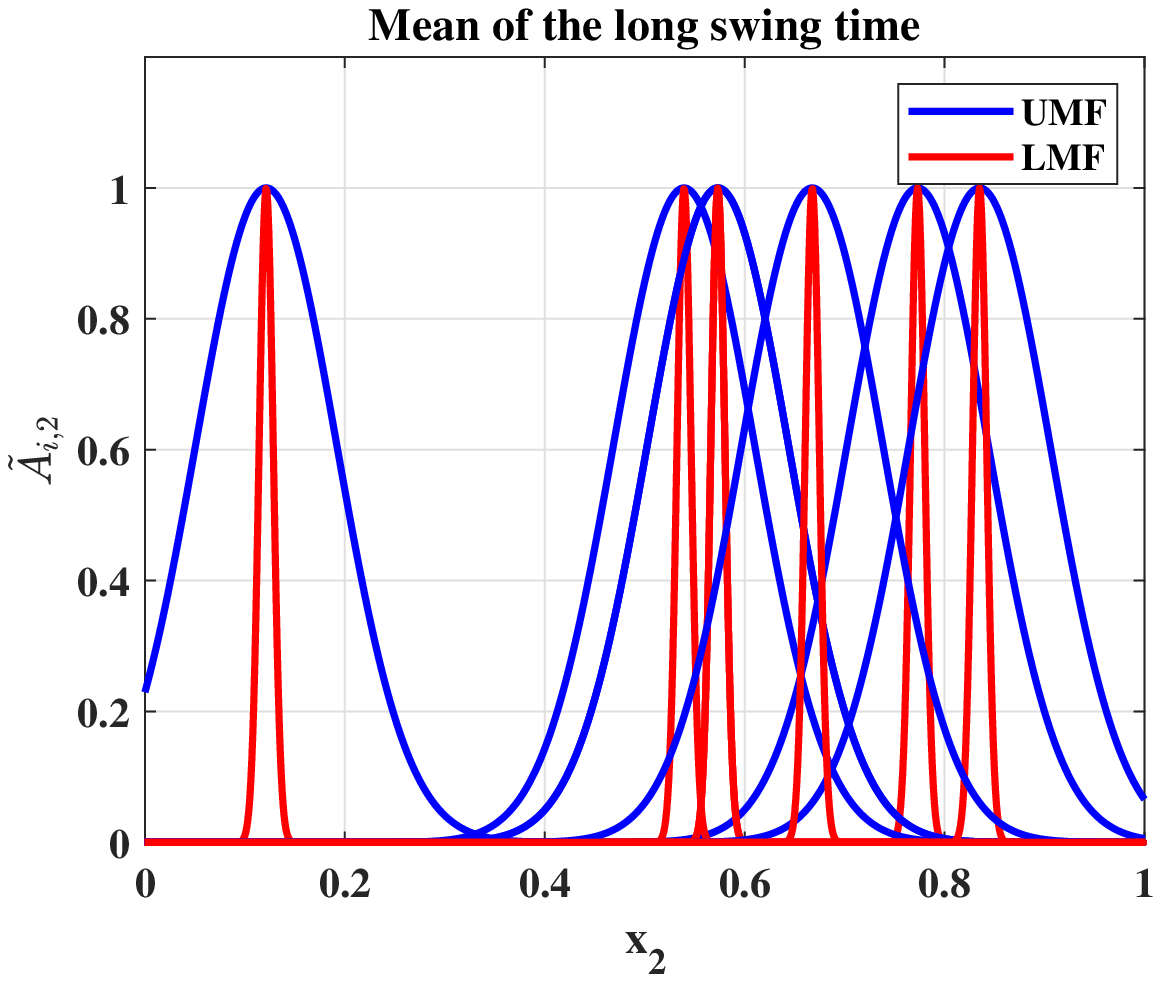}} \\
    \subfigure[][]{\includegraphics[width = 1.8in]{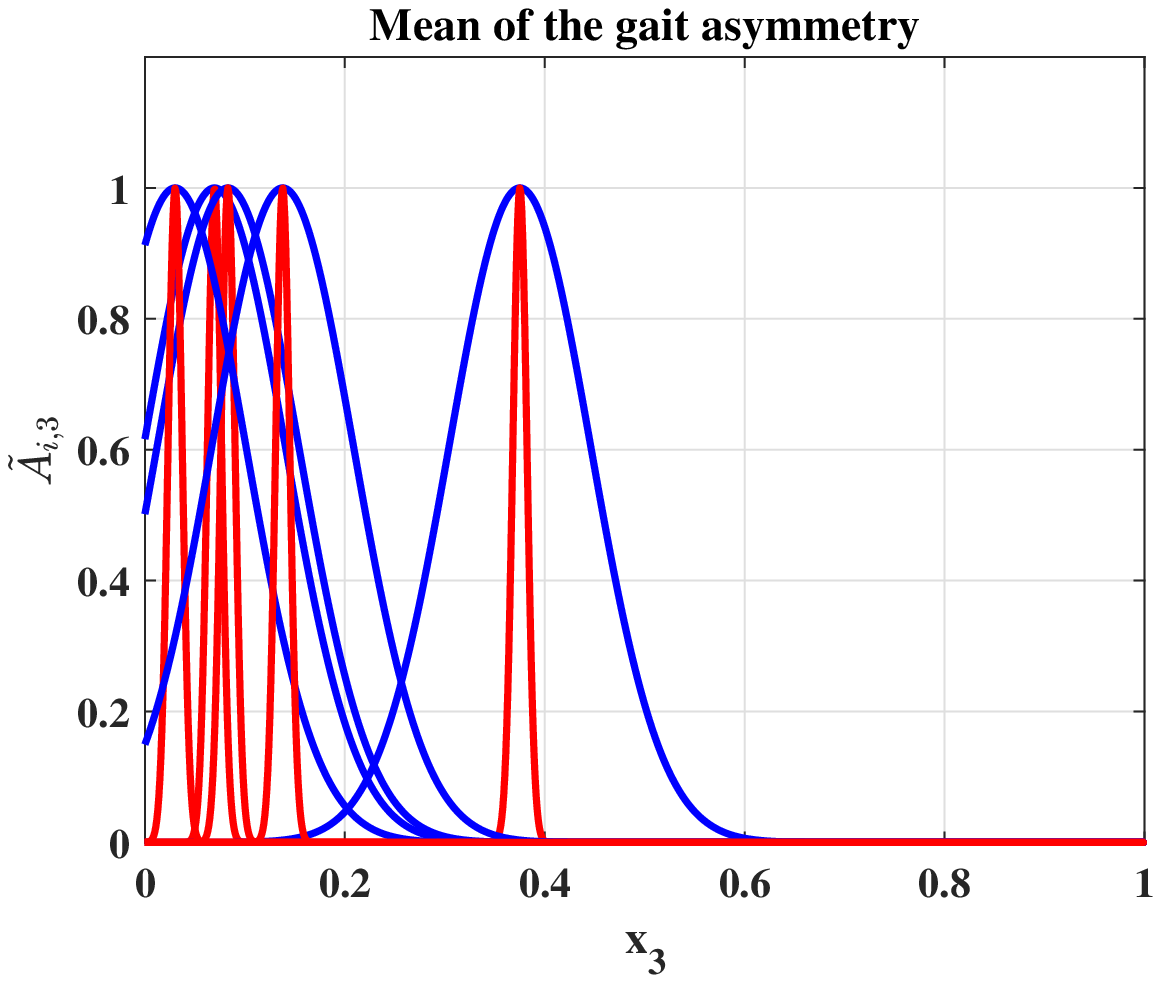}} &
    \subfigure[][]{\includegraphics[width = 1.8in]{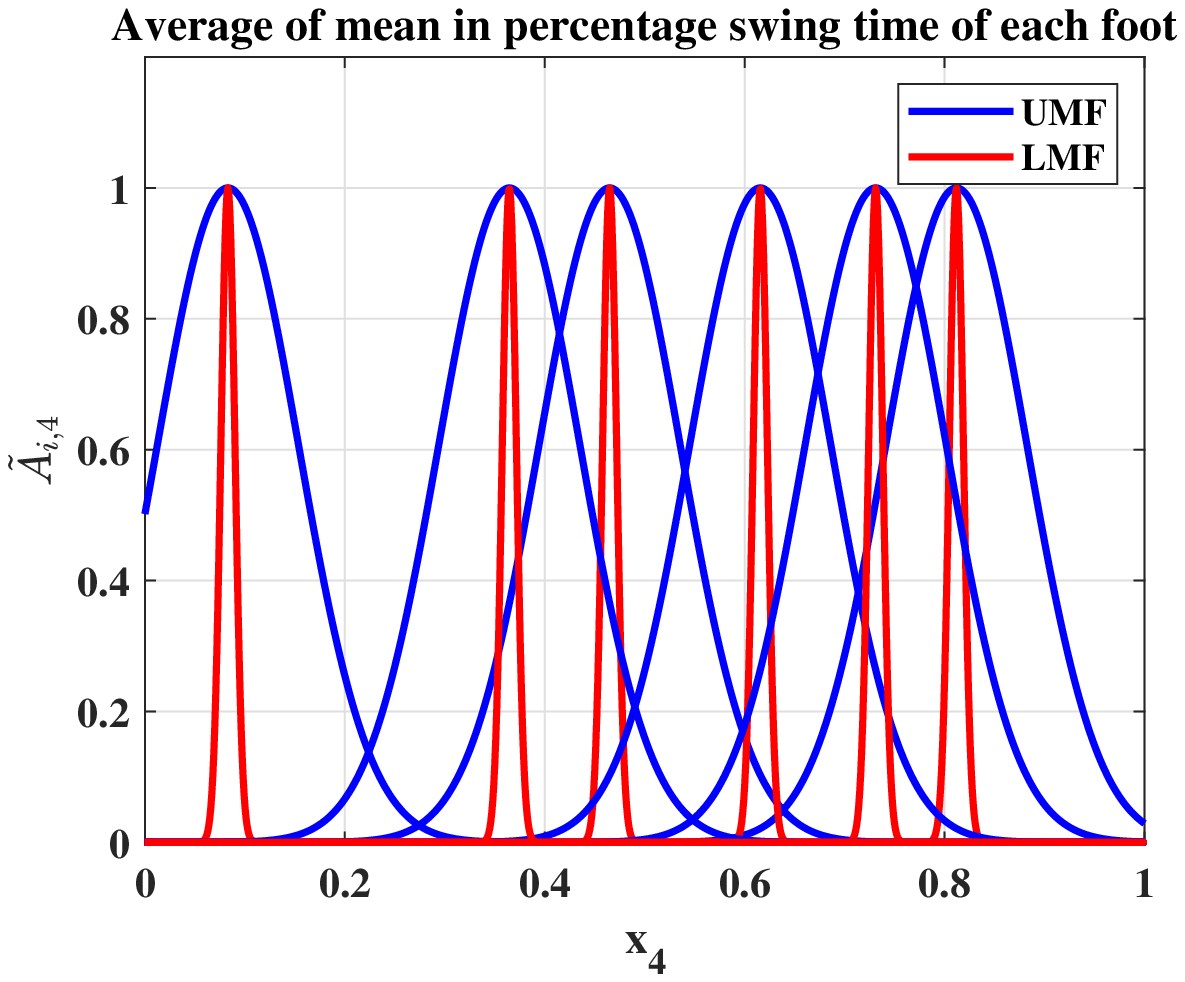}} \\
    \subfigure[][]{\includegraphics[width = 1.8in]{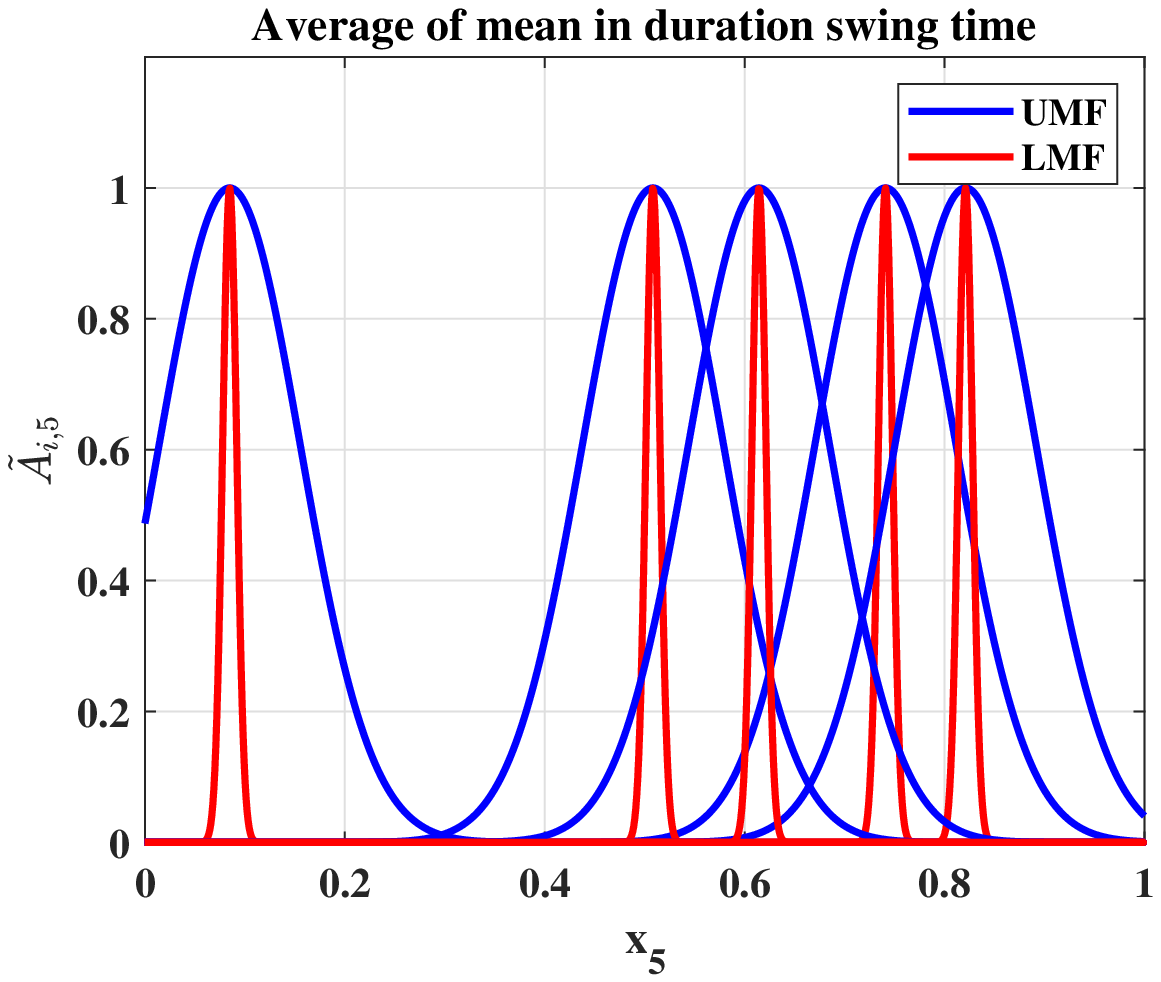}} &
    \subfigure[][]{\includegraphics[width = 1.8in]{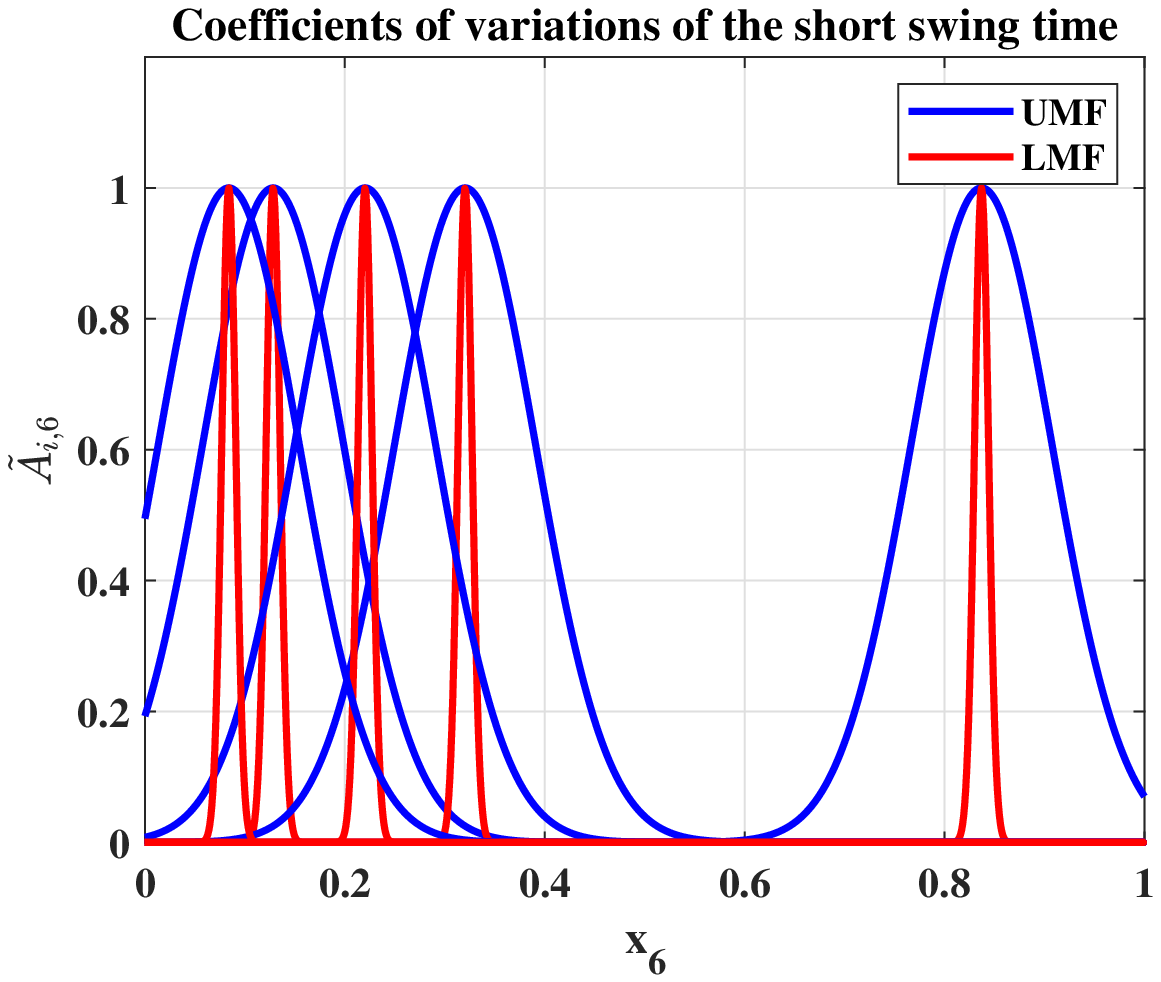}} \\
    \subfigure[][]{\includegraphics[width = 1.8in]{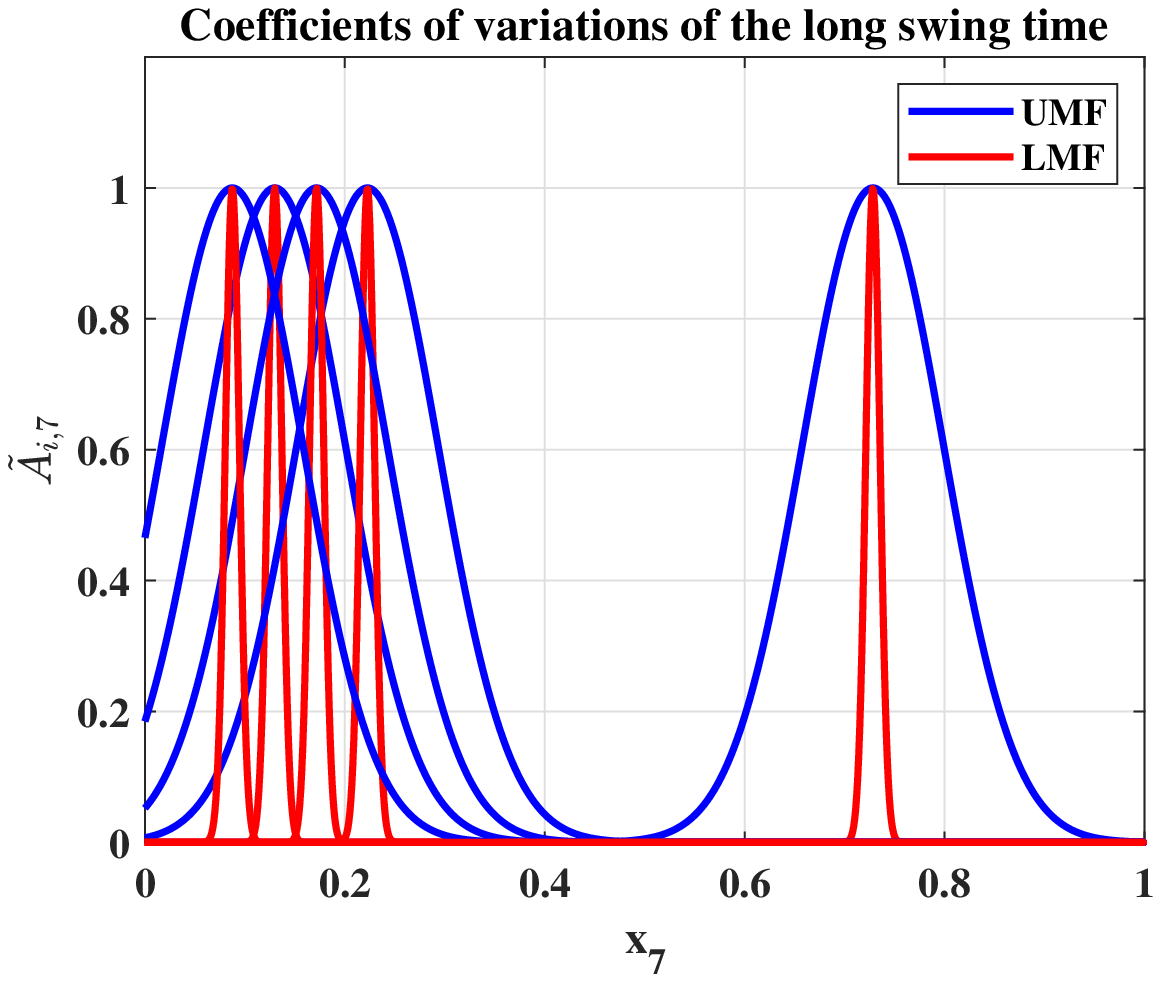}} &
    \subfigure[][]{\includegraphics[width = 1.8in]{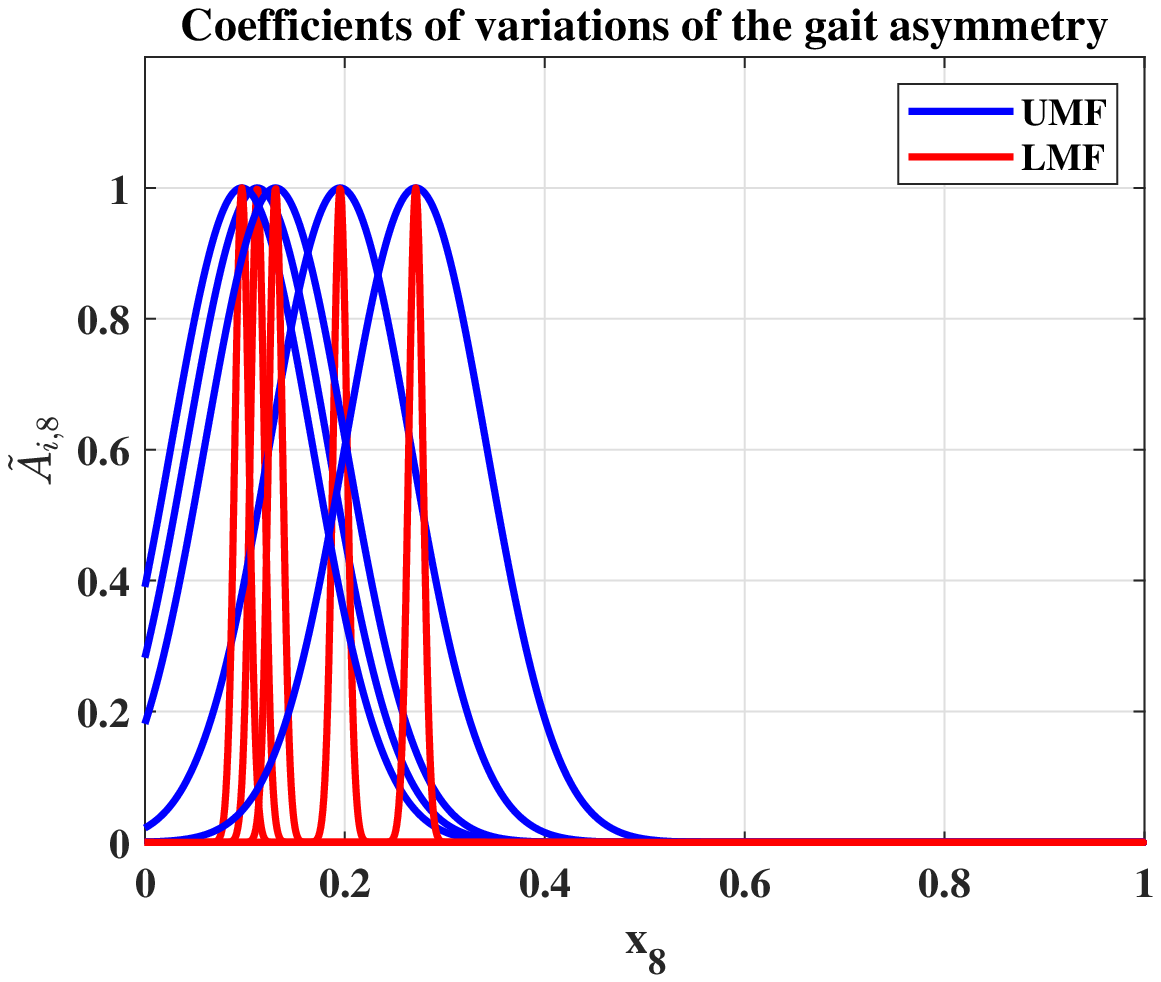}} \\
    \subfigure[][]{\includegraphics[width = 1.8in]{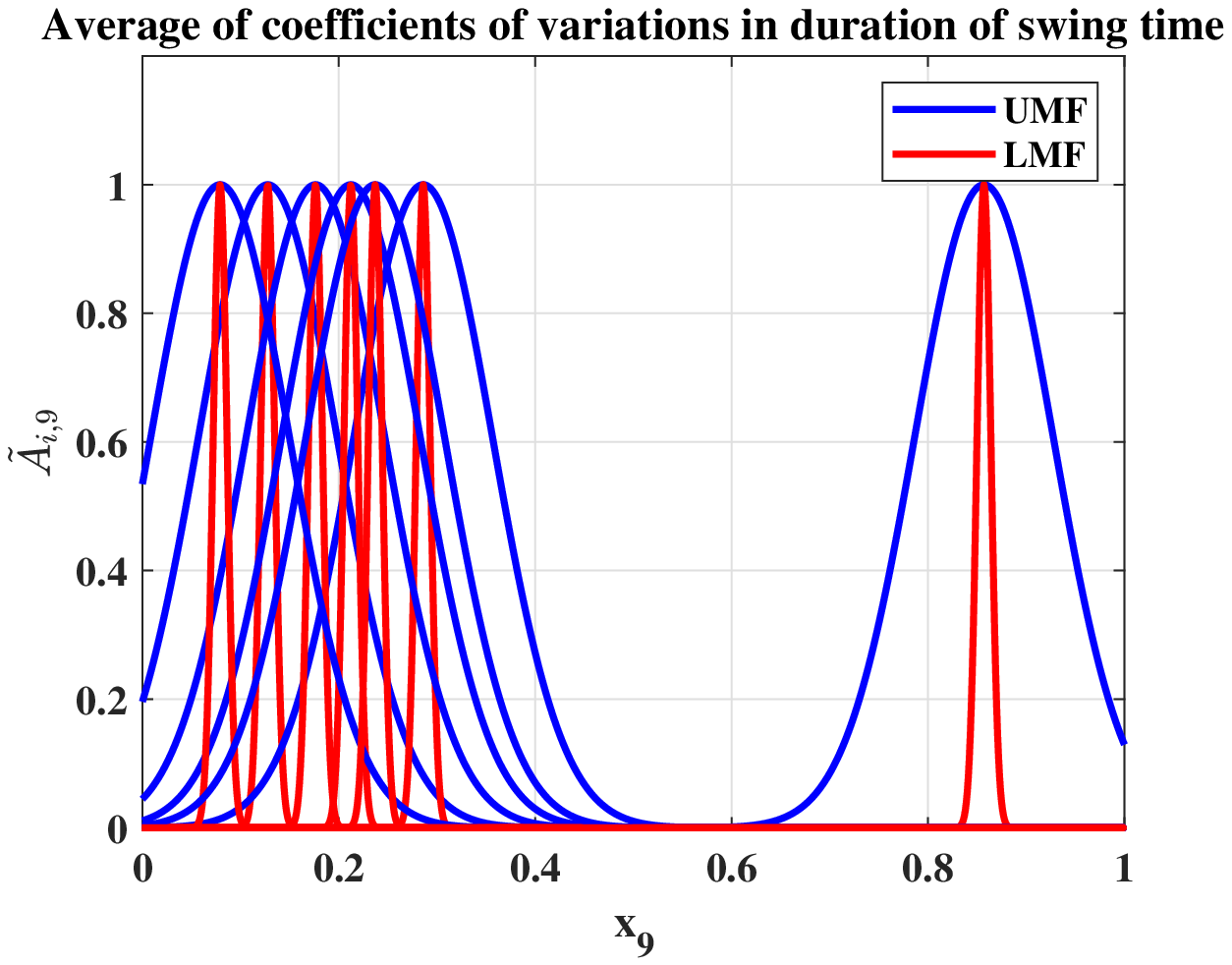}} &
    \subfigure[][]{\includegraphics[width = 1.8in]{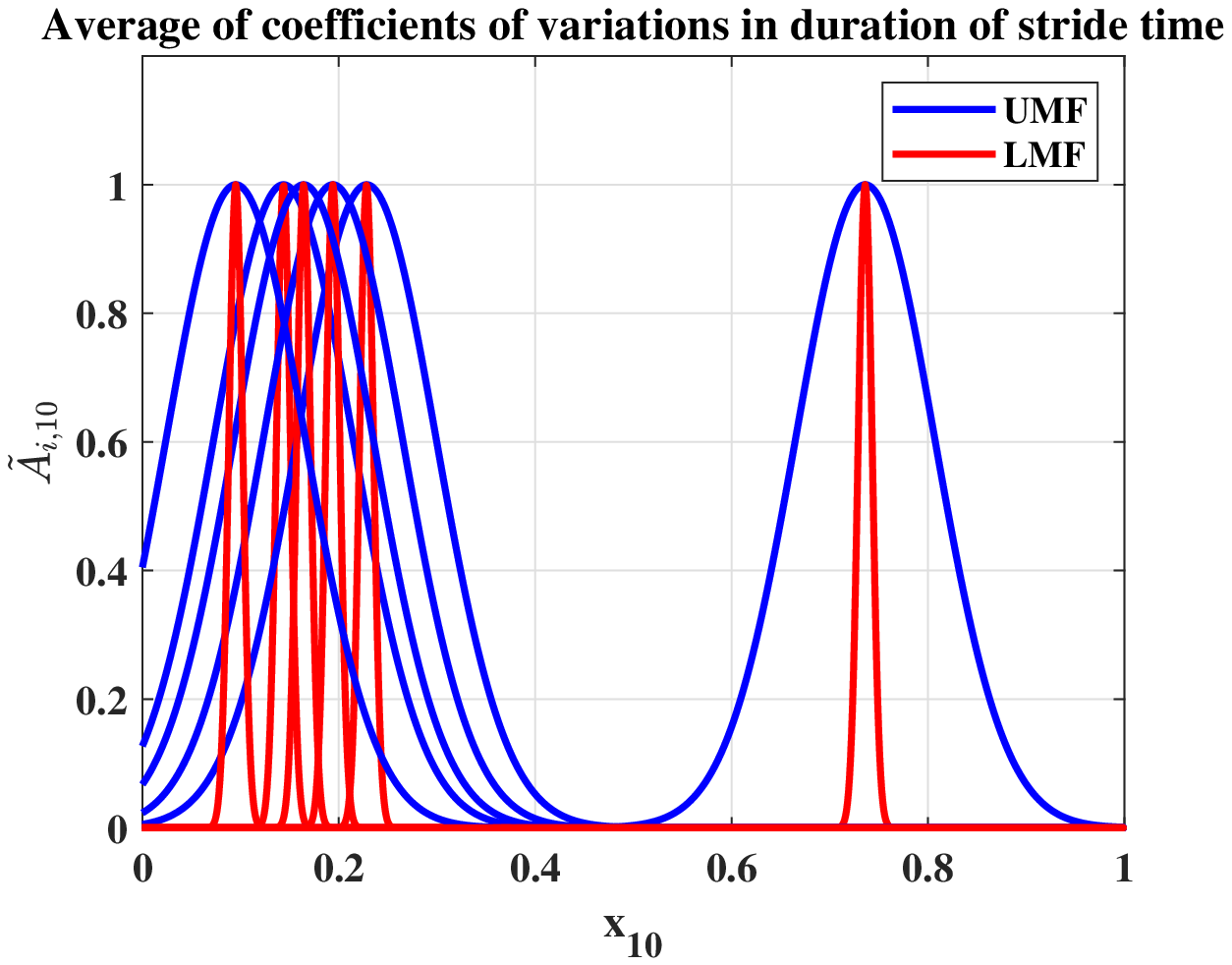}} \\
\end{tabular}
\caption{Extracted Interval Type-2 fuzzy sets for different features.}
\label{fig_fuzzy_sets}
\end{figure}

\subsection{Final Extracted Fuzzy Rules}
\begin{figure}[h]
\centering
    \includegraphics[width = 5.5in]{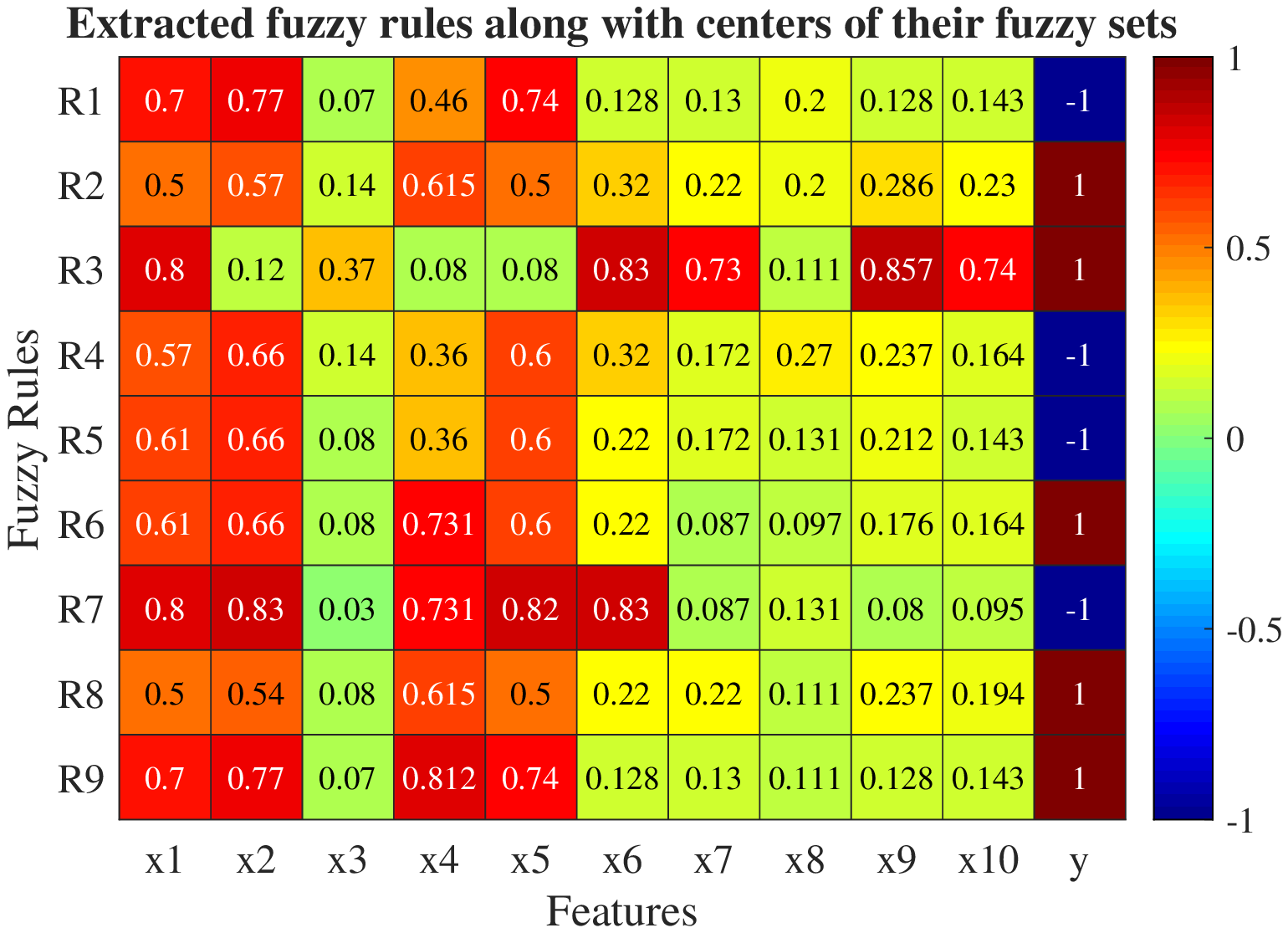}
\caption{Extracted fuzzy rules. Each row indicates a fuzzy rule ($R_1$ to $R_9$) and each column except the last indicates a feature ($x_1$ to $x_{10}$). The last column presents each fuzzy rule's output value as the consequent part's parameter ('+1' for patients (colored red) and '-1' for healthy subjects (colored blue). The center of fuzzy sets used for constructing each rule is shown in each cell and the color of each cell represents the sense of the linguistic variable (higher values are near to red and lower values are close to white).}
\label{fig_rules}
\end{figure}

In this part, whole available samples are fed to the Interval Type-2 Fuzzy Neural Network to extract final fuzzy rules based on the currently available data. The number of fuzzy rules is set to 9. The performance of the method is summarized in Table \ref{table_final}. Figure \ref{fig_fuzzy_sets} shows extracted fuzzy sets for all features. It is obvious that based on the available data and importance of different ranges of values, the density of defined fuzzy sets in different regions of different features is various. Figure \ref{fig_rules} shows the extracted fuzzy rules. Each row indicates a fuzzy rule and each column indicates a feature. The center of fuzzy sets used for constructing each rule is shown in each cell and the color of each cell represents the sense of the linguistic variable (higher values are near to red and lower values are close to white). The last column shows the consequent parts' parameters which is the proposed label of each fuzzy rule ('+1' for patients (colored red) and '-1' for healthy subjects (colored blue).

\section{Conclusions}
\label{sec6}
In this paper, a classifier using an interval type-2 fuzzy neural network is proposed to diagnose patients suffering from Parkinson's Disease (PD). The proposed method analyzes the gait cycle of subjects recorded in the form of vertical Ground Reaction Force (vGRF) signals. To have an interpretable system, 10 clinical features are extracted from the vGRF signal. To overcome the uncertainty and sensor noisy measurements, the fuzzy neural network is built upon interval type-2 fuzzy logic.

To extract fuzzy rules, two paradigms are presented. First, using the available training dataset, initial interpretable fuzzy rules are extracted. To able system learning from new instances, an online learning approach is proposed. This complementary online learning investigates the performance and coverage of the current rule base. In the case of wrong classification and low coverage, a new fuzzy rule is added to the rule base.

The performance of the proposed method is evaluated on three datasets obtained from different subjects (\cite{yogev2005dual,frenkel2005treadmill,hausdorff2007rhythmic}) and compared with other supervised and unsupervised machine learning approaches using similar clinical features. Moreover, the performance of the model in the presence of noise is evaluated and compared with a T1 fuzzy neural network. The effectiveness of online learning is investigated by adding new training instances after finishing batch learning. This experiment shows an obvious improvement in the network's performance after applying the proposed online learning. Finally, the extracted fuzzy sets of different features are reported.

Using meta-heuristic optimization methods like Evolutionary Algorithms and Swarm Intelligence method for fine-tuning rules' parameters is proposed as the future study.

\clearpage
\bibliographystyle{plain}
\bibliography{myref}

\end{document}